%% file: main.tex
\documentclass[10pt,twocolumn,letterpaper]{article}

\usepackage[pagenumbers]{cvpr} 

\input{preamble}

\definecolor{cvprblue}{rgb}{0.21,0.49,0.74}
\usepackage[pagebackref,breaklinks,colorlinks,citecolor=cvprblue]{hyperref}
\usepackage[capitalize]{cleveref}

\def\paperID{11893} 
\def\confName{CVPR}
\def\confYear{2024}

\title{vid-TLDR: Training Free Token merging for \\Light-weight Video Transformer}

\newcommand*\samethanks[1][\value{footnote}]{\footnotemark[#1]}
\author{\textbf{Joonmyung Choi}\thanks{equal contribution.} \hspace{0.4cm} \textbf{Sanghyeok Lee}\samethanks \hspace{0.4cm} \textbf{Jaewon Chu} \hspace{0.4cm} \textbf{Minhyuk Choi} \hspace{0.4cm} \textbf{Hyunwoo J. Kim}\thanks{Corresponding author.} \vspace{0.4cm}
\\
Department of Computer Science and Engineering, Korea University\\
{\tt\small \{\href{mailto:pizard@korea.ac.kr}{pizard}, \href{mailto:cat0626@korea.ac.kr}{cat0626}, \href{mailto:allonsy07@korea.ac.kr}{allonsy07}, \href{mailto:sodlqnf123@korea.ac.kr}{sodlqnf123}, \href{mailto:hyunwoojkim@korea.ac.kr}{hyunwoojkim}\}@korea.ac.kr}
}

\begin{document}
\twocolumn[{
\maketitle
\input{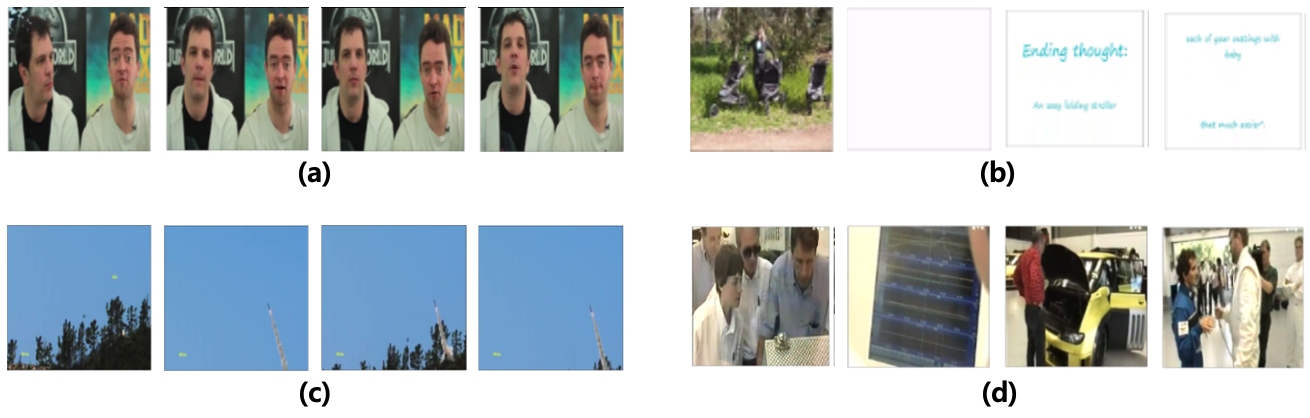}
}]
\input{section/abstract}
\input{section/introduction}
\input{section/preliminaries}

\input{section/method}

\input{section/experiments}

\input{section/relatedworks}
\input{section/conclusion}
{
    \small
    \bibliographystyle{ieeenat_fullname}
    \bibliography{main}
}

\maketitlesupplementary
\setcounter{table}{0}
\renewcommand{\thetable}{\Alph{table}}
\setcounter{figure}{0}
\renewcommand{\thefigure}{\Alph{figure}}
\appendix
\section{A deeper exploration into temporal bias} We here present the discussion for the behavior of video Transformers regarding temporal bias. We have observed that the positional encoding in temporal-spatial attention even has learned the data bias, \ie, the main contents of the clip are generally positioned at the front or middle. For better understanding, we visualize the attention map built with 1) positional encoding (PE) + representation, 2) representation only, and 3) PE only. \Cref{fig:att} reveals that PE of video Transformers causes the temporal bias, \ie focusing on the early frame.
\input{rebuttal/fig_att}

\section{Detailed reduced number of tokens}
In this section, we present the hyperparameters indicating the number of reduced tokens for each dataset and task, as shown in \Cref{sup:reduced_r}. According to the goal of vid-TLDR, which focuses on early-stage merging, all experiments involved reduction in the first four layers.
\input{Supple/reduced_r}

\section{Detailed video-text retrieval}
In Table 2 of the main paper, we provide video-text retrieval by averaging the performances of video-to-text retrieval and text-to-retrieval. Here, we provide the detailed results of each task in \Cref{tab:retrieval_t2v} with MSRVTT~\cite{xu2016msr}, MSVD~\cite{chen-dolan-2011-collecting}, ActivityNet~\cite{caba2015activitynet}, DiDeMo~\cite{anne2017localizing}, LSMDC~\cite{rohrbach2017movie}, SSV2-Label/Template~\cite{lei2022revealing}.
\input{Tables/retrieval_t2v}

\section{Zero-shot retrieval results}
To validate the effectiveness of vid-TLDR, we report additional zero-shot retrieval results for baseline model~\cite{li2023unmasked}, existing token merging method (ToMe~\cite{bolya2022token}) and ours across MSVD~\cite{chen-dolan-2011-collecting}, MSRVTT~\cite{xu2016msr}, DiDeMo~\cite{anne2017localizing}, ActivityNet~\cite{caba2015activitynet}. All experiments were conducted with evaluation only, without additional training. The average results for text-to-video and video-to-text retrieval accuracy across these four datasets are presented in Table \ref{sup:zero-shot}, \ref{sup:zs_msvd}, \ref{sup:zs_msrvtt}, \ref{sup:zs_didemo}, and \ref{sup:zs_anet}.
We conducted a grid search to determine the effective number of tokens to reduce.
In the case of ToMe, tokens were merged uniformly across layers following the default configuration, and a grid search was conducted with 20-25 tokens per layer.
On the other hand, our method merged in the first four front layers, exploring total reductions of 300, 330, and 360 tokens.
Across all configurations, vid-TLDR consistently achieves higher R@1 with lower FLOPs compared to ToMe.
On average, it achieves 25.0\% and 11.8\% reduction in FLOPs, along with 2.6\% and 2.2\% higher R@1 for UMT-B and UMT-L.



\input{Supple/zero_shot_mean}

\input{Supple/zero_shot_1}
\input{Supple/zero_shot_2}


\end{document}

%% file: preamble.tex
\usepackage[dvipsnames]{xcolor}

\usepackage{url}

\usepackage{booktabs}       
\usepackage{amsfonts}       
\usepackage{nicefrac}       
\usepackage{microtype}      
\usepackage{colortbl}         

\usepackage{graphicx}
\usepackage{multirow}
\usepackage{amsmath}
\usepackage{subcaption}
\usepackage{graphicx}
\usepackage{algorithm}
\usepackage{algpseudocode}

\definecolor{Gray}{rgb}{0.8,0.8,0.8} 
\definecolor{LightBlue}{rgb}{0.8,0.9,1} 
\definecolor{LightOrange}{rgb}{1,0.9,0.8} 
\definecolor{Cyan}{rgb}{0,1,1}

\makeatletter
\def\thickhline{%
  \noalign{\ifnum0=`}\fi\hrule \@height \thickarrayrulewidth \futurelet
   \reserved@a\@xthickhline}
\def\@xthickhline{\ifx\reserved@a\thickhline
               \vskip\doublerulesep
               \vskip-\thickarrayrulewidth
             \fi
      \ifnum0=`{\fi}}
\makeatother

\newlength{\thickarrayrulewidth}

%% file: Figures/Fig1.tex
\begin{center}
     \centering
     \captionsetup{type=figure}
     \includegraphics[width=0.8\textwidth, trim=0 0 0 0, clip]{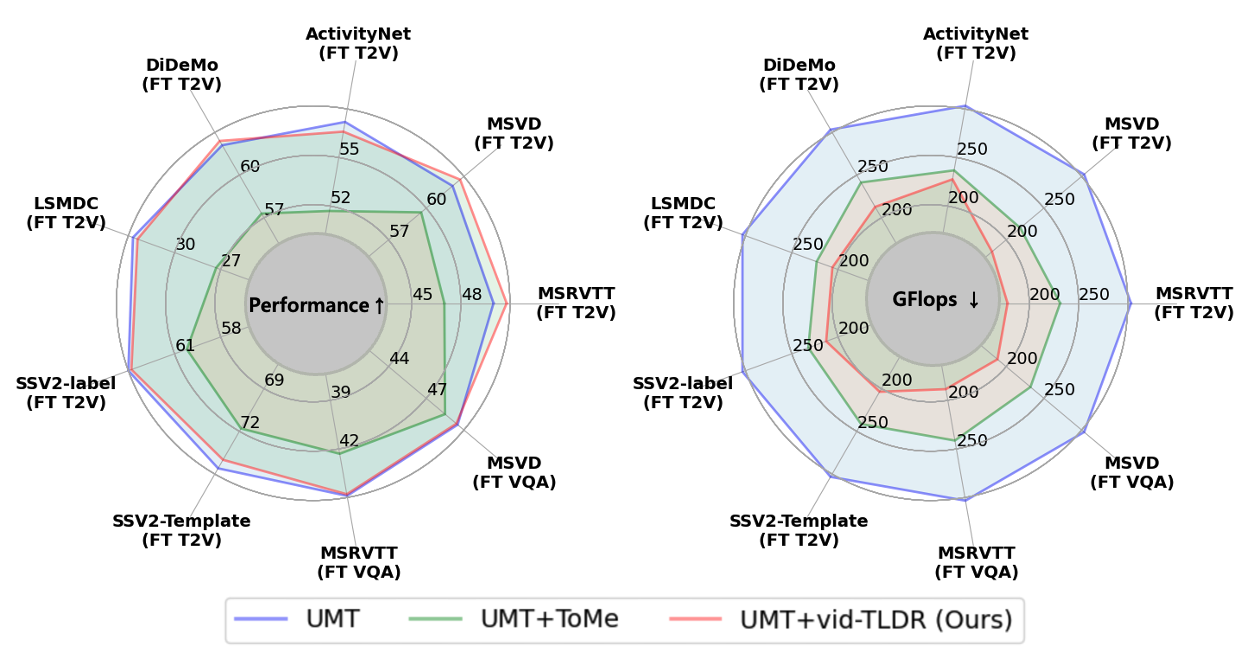}
     \captionof{figure}{
     Comparison of vid-TLDR (Ours) with UMT~\cite{li2023unmasked}. Without any additional training, vid-TLDR obtains comparable or even better performance than the base model UMT (left) while reducing the considerable computational cost (right). UMT-B (87M) is used.}
     \label{fig:1}
\end{center}

%% file: section/abstract.tex
\begin{abstract}
Video Transformers have become the prevalent solution for various video downstream tasks with superior expressive power and flexibility. 
However, these video transformers suffer from heavy computational costs induced by the massive number of tokens across the entire video frames, which has been the major barrier to train and deploy the model.
Further, the patches irrelevant to the main contents, \eg, backgrounds, degrade the generalization performance of models.
To tackle these issues, we propose training-free token merging for lightweight video Transformer (vid-TLDR) that aims to enhance the efficiency of video Transformers by merging the background tokens without additional training.
For vid-TLDR, we introduce a novel approach to capture the salient regions in videos only with the attention map.
Further, we introduce the saliency-aware token merging strategy by dropping the background tokens and sharpening the object scores.
Our experiments show that vid-TLDR significantly mitigates the computational complexity of video Transformers while achieving competitive performance compared to the base model without vid-TLDR.
Code is available at \url{https://github.com/mlvlab/vid-TLDR}.
\end{abstract}

%% file: section/introduction.tex
\section{Introduction}
\label{sec:1}
With the success of Transformers in computer vision, \eg, classification~\cite{dosovitskiy2020image,touvron2021training}, object detection~\cite{carion2020end,wang2022anchor,meng2021-CondDETR,zhu2020deformable,li2022dn,zang2022open}, segmentation~\cite{wang2020end,xie2021segformer}, a line of works~\cite{fu2021violet,tong2022videomae,wang2023videomae,li2023unmasked,wang2022internvideo,zellers2021merlot} have proposed video Transformers to comprehend the video for various downstream tasks.
The attention mechanism in Transformers shows the desirable characteristics for video understanding such as the ability to capture the spatial and temporal dependencies at the same time.
Consequently, these video Transformers have been the primary backbones for the various downstream tasks in the video domain, including action recognition~\cite{yang2022recurring, xing2023svformer}, video-text retrieval~\cite{liu2021hit, gabeur2020multi}, video question-answering~\cite{xiao2022video, gao2023mist}, etc.
Meanwhile, the self-attention mechanism entails the dot-product calculation between tokens, which brings the quadratic cost in the number of tokens.
This poses a challenge for existing video Transformers like UMT~\cite{li2023unmasked} that tokenize the whole video into a large number of tokens.\\
\noindent In the image domain, several works have tried to mitigate the heavy computation of attention by refining the attention itself~\cite{wang2020linformer, kitaev2020reformer, beltagy2020longformer, xiong2021nystromformer}, or limiting the range of attention by a pre-defined window~\cite{liu2021swin, chu2021twins}.
Yet, these works are not a favorable solution for video Transformers since
the methods entail architectural changes, requiring re-training video models with large datasets.
As an alternative, in the image domain several works have proposed `training-free' token reduction methods using the flexibility of Transformers in handling a variable number of input tokens.
For instance, prior works~\cite{liang2022evit,rao2021dynamicvit} simply prune or merge the uninformative tokens to reduce the computational cost based on attentiveness.
However, we observed that the existing training-free token reduction methods for images are suboptimal for video transformers.
First, previous attention-based informative scores are not accurate enough to use in early layers as discussed in~\cite{liang2022evit}. 
So, the token reduction cannot be performed at earlier layers.
Further, the attention scores of video Transformers contain a temporal bias, which makes it difficult to directly adopt them as the informativeness of the tokens, see~\Cref{fig:2}.\\
\noindent Based on these observations, we propose vid-TLDR, \textbf{T}raining-free token merging for \textbf{L}ight-weight vi\textbf{D}eo TransformeR, to effectively merge the tokens through two steps. 
First, we conduct \textit{saliency detection via attention sharpness}.
We observe that our proposed metric understands the salient region, which is more informative than backgrounds, even with the attention map in the first layer of Transformers. 
We also introduce the \textit{saliency-aware token merging}, a training-free plug-in module to suppress the tokens irrelevant to the target tasks. 
Through saliency-aware token merging, we effectively drop the information of tokens in backgrounds and further contrast the informativeness of the foreground objects. 
Based on these components, we minimize the hindrance by irrelevant tokens from the early layers of video Transformers.
Through experiments, we show that, without any additional training, the adoption of vid-TLDR brings performance improvements of (+0.8\%, +0.5\%, +1.1\%) with at least 39.5\% lower FLOPs in UMT-B~\cite{li2023unmasked} on MSRVTT~\cite{xu2016msr}, MSVD~\cite{chen-dolan-2011-collecting}, DiDeMo~\cite{anne2017localizing}, respectively.
To summarize, the contributions of vid-TLDR are presented as follows:
\begin{itemize}
    \item We propose the novel token merging method vid-TLDR, which reduces the tokens irrelevant to target tasks from the early layers of the video Transformer.
    \item We detect the salient region of videos based on the sharpness of the attention scores even from the first layer. 
    \item Based on the saliency scores, we also propose saliency-aware token merging with the masked saliency scores for adaptively adjusting the informativeness of the tokens. 
    \item vid-TLDR shows the competitive performance with the baselines across four benchmarks in video-text retrieval and two benchmarks in video question-answering. It is worth noting that vid-TLDR even shows superior performance while reducing the computational complexity.
\end{itemize}

%% file: section/preliminaries.tex
\input{Figures/Fig2}

\section{Preliminaries}
\label{sec:2}
In this section, we briefly review the video Transformers and token reduction approaches, then introduce techniques to measure the informativeness of tokens using the attention map of the video Transformers.

\noindent\textbf{Video Transformer.}
In Transformers~\cite{vaswani2017attention}, the self-attention mechanism is defined as 
\begin{align}
    \text{Attention}(Q,K,V) = \text{softmax}\left (\frac{QK^\top}{\sqrt{C}} \right )V,
\end{align}
where $Q,K,V \in \mathbb{R}^{N \times C}$ are the projection of the tokens $X\in \mathbb{R}^{N \times C}$ by learnable matrices $W_Q,W_K,W_V \in \mathbb{R}^{C \times C}$. 
Given a video clip composed of $T$ frames in the resolution of $H \times W$, video Transformers first generate the tokens $X\in\mathbb{R}^{N \times C}$ by considering the clip as the set of tubes, where $N = \frac{T}{t} \times \frac{H}{P} \times \frac{W}{P}$, and ($t \times P \times P)$ is the size of each tube. 
All tokens in the video Transformers interact with others across the frames by spatio-temporal attention. 
Despite the advantage of capturing both spatial and temporal dependencies, it also demands enormous computational resources to handle the large number of tokens.
Compared to one image, the computational cost for a clip is increased by $(\frac{T}{t})^2$ times since the cost of attention is quadratic in the number of tokens.
This cost further increases as the number of frames per tube $t$ decreases.

\noindent\textbf{Token Reduction Methods.}
\label{2.2}
Based on the flexibility of Transformers in the number of tokens, token reduction approaches~\cite{liang2022not,rao2021dynamicvit,bolya2022token,kong2022spvit, fayyaz2022adaptive} reduce the intermediate tokens by pruning or merging them, leading to the lower computational cost $O((N^\prime)^2C + N^\prime C^2)$, where $N^\prime < N$. 
To minimize information loss after reduction, they mainly prune/merge the tokens based on the attentiveness defined as 
\begin{align}
a_\text{cls} = \text{softmax} \left(\frac{q_\text{cls} K^\top}{\sqrt{C}}\right),
\label{eq:2}
\end{align}
where $q_\text{cls}$ is the query vector of the class token.
Prior works achieve a competitive performance with the original model through additional training.
In parallel, ToMe~\cite{bolya2022token} has demonstrated the possibility of training-free token merging in the image domain using the similarity of the tokens.
This simple plug-in approach is favorable for video Transformers considering its huge complexity, yet the capability to capture the informativeness of the tokens is absent.

\noindent\textbf{Informativness of tokens.}
Previous works~\cite{rao2021dynamicvit,liang2022evit}, due to the low reliability of the attention map in the early stage, could not reduce the tokens in the earlier layers.
However, we believe that the early pruning/merging is desirable for video Transformers from two perspectives: 1) it prevents the interaction between the tokens irrelevant to the main contents by only retaining the salient tokens, and 2) it largely alleviates the complexity of the entire layers with fewer tokens from the beginning of Transformer.
To validate this, we explore whether the attention map is a sufficient approximation of the informativeness estimator even in the first layer.
Note that, due to the absence of the class tokens in recent video Transformers, we modify \Cref{eq:2} by summarizing the whole query vector as
\begin{align}
\bar{a} = \frac{1}{N}{\sum}^N_i A_i,
\label{eq:3}
\end{align}
where $A_i$ is the $i$-th row vector of the attention matrix $A=\text{softmax}\left(\frac{Q K^\top}{\sqrt{C}}\right)$.
Further, we visualize the attention rollout~\cite{abnar2020quantifying}, which is the well-known saliency detector by quantifying the flow of attention from the tokens in the $l$-th layer to output, defined as
\begin{align}
\tilde{A}^l &= {\prod^\curvearrowleft}^L_{i=l} A^i \text{ and } \tilde{a}^l = \frac{1}{N}{\sum}^N_i \tilde{A}^l_i,
 \label{eq:4}
\end{align}
where $L$ is the number of layer in Transformers, and $A^i\in \mathbb{R}^{N \times N}$ is the attention map in $i$-th layer.
As shown in~\Cref{fig:2}, the attentiveness $\bar{a}$ failed to capture salient tokens due to the low reliability of the attention map in the early stage, and attention rollout $\tilde{a}$ also largely confused the foreground objects.
Further, we have observed the temporal biases of video Transformers, \eg, the later frames exhibit lower activation regardless of the importance of the frames.


%% file: Figures/Fig2.tex
\begin{figure*}[t!]
     \centering
     \begin{subfigure}[b]{0.9\linewidth}
         \centering
         \includegraphics[trim=0 0 0 0,clip, width=\linewidth]{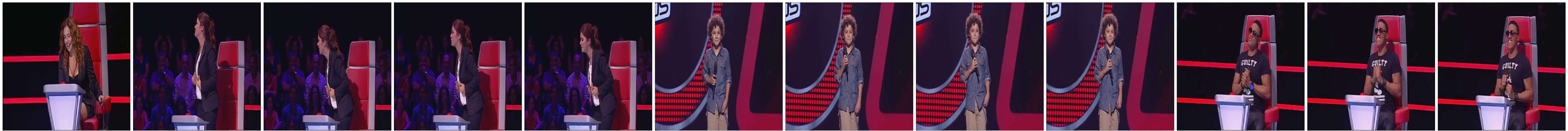}
         \caption{Input}
         \label{fig:2.1}
     \end{subfigure}
     \hfill
     \begin{subfigure}[b]{0.9\linewidth}
         \centering
         \includegraphics[trim=0 0 0 0,clip, width=\linewidth]{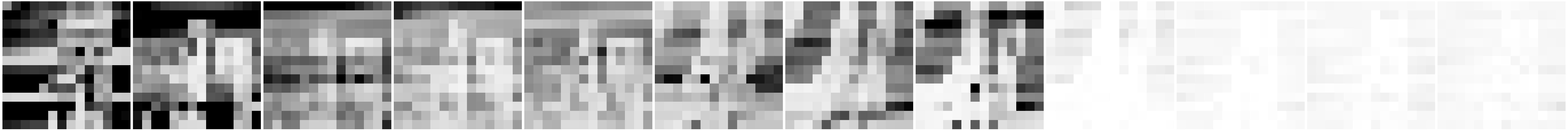}
         \caption{Attentiveness $\bar{a}$ in~\Cref{eq:3}}
         \label{fig:2.2}
     \end{subfigure}
     \begin{subfigure}[b]{0.9\linewidth}
         \centering
         \includegraphics[trim=0 0 0 0,clip, width=\linewidth]{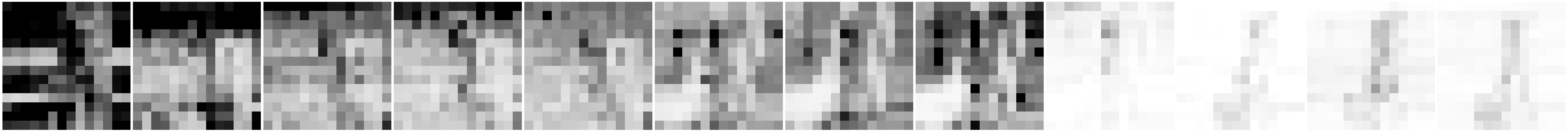}
         \caption{Attention Rollout $\tilde{a}$ in~\Cref{eq:4}}
         \label{fig:2.3}
     \end{subfigure}
     \hfill
     \begin{subfigure}[b]{0.9\linewidth}
         \centering
         \includegraphics[trim=0 0 0 0,clip, width=\linewidth]{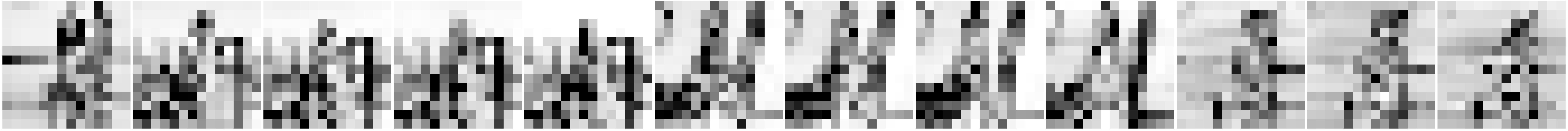}
         \caption{Ours $s$ in~\Cref{eq:5}}
         \label{fig:2.4}
     \end{subfigure}
     \vspace{-5pt}
     \caption{Visualization of the attention map of each method in the first layer. Both attentiveness $\bar{a}$ and attention Rollout $\tilde{a}$ confused with foreground objects and the background, also have a temporal bias, resulting in overall low attention in the later frames. These problems are mitigated in our method, focusing on the object across all frames.}
     \label{fig:2}
\vspace{-10pt}
\end{figure*}

%% file: section/method.tex
\section{Method}
\input{Figures/Fig_main}

\input{Figures/Fig_sharpness}

\label{sec:3}
We introduce vid-TLDR, \textbf{T}raining free token merging for \textbf{L}ight-weight vi\textbf{D}eo transforme\textbf{R}. The goal of vid-TLDR is to effectively merge the tokens from the early stage by two steps: 1) Saliency detection via attention sharpness (\Cref{sec:3.1}), 2) Saliency-aware token merging (\Cref{sec:3.2}).

\subsection{Saliency detection via attention sharpness}
\label{sec:3.1}
As discussed in~\Cref{sec:2}, existing works do not reduce the tokens in the first few layers because of the low reliability. However, we believe that tokens irrelevant to the target tasks should be reduced as early as possible to minimize their adverse influence. To this end, we analyze the attention scores $A_i = \text{softmax}(\frac{q_i K^\top}{\sqrt{C}})$ in the first self-attention layer concerning the foreground and background tokens.
\Cref{fig:sharp} reveals that the background tokens are quite equally affected by neighboring tokens, whereas the tokens of foreground objects gather the information from more specific tokens showing sharper attention scores compared to backgrounds.
Based on this observation, we devise a \textit{sharpness function} $S$ to capture the saliency of tokens with entropy. 
Specifically, using the negative entropy given as $H_i = {\sum}^N_j A_{ij} \log A_{ij}$, we define the sharpness function $S$ as
\begin{align}
s_i=S(H_i) = \frac{H_i - \min(H)}{\max(H) - \min(H)},
\label{eq:5}
\end{align}
where $H = [H_1,\ldots,H_N]$. 
Since an informative foreground token has low entropy, its saliency score $s_i$ is usually high. 

\noindent\textbf{Remarks.} 
To validate the saliency scores $s$, we conduct the experiments by pruning the 400 tokens in each layer of UMT~\cite{li2023unmasked} on video-text-retrieval task with MSRVTT~\cite{xu2016msr}.
In \Cref{table:merging}, we demonstrate that the pruning by the proposed score accelerates the base model (first row) with the competitive performance across layers.
Interestingly, the token reduction in the earlier layers is more effective.
To be specific, the token reduction in the first layer shows the best results (50.4\%) with the lowest FLOPs (237.6 (G)).
Based on this observation, we mainly applied our token reduction to earlier layers.
Qualitative results in~\Cref{fig:2.4} show that the saliency scores in the first layer successfully detect the salient region. 


\input{Tables/merging}

\input{Tables/retrieval}
\subsection{Saliency-aware token merging}
\label{sec:3.2}
Given the saliency scores $s = [s_1,...,s_N]$ of the tokens, our goal is to merge tokens while 
suppressing the influence of the irrelevant tokens with low saliency scores.
We start with a brief review of a training-free token reduction method, ToMe~\cite{bolya2022token}. 
ToMe splits tokens $X$ into two sets $X^\text{src}, X^\text{dst} \subset X$ and performs the bipartite soft matching between the two sets
to form token groups. 
For each group $\mathcal{G}_i$, the features are aggregated as 
\begin{align}
x^\prime_i = {\sum}_{j\in{\mathcal{G}_i}} \frac{m_j x_j}{{\sum}_{j^\prime \in{\mathcal{G}_i}} m_{j^\prime} }, \label{eq:6}
\end{align}
where $\{x_i\}_{i \in \mathcal{G}_i} \subset X$, $\forall_{\mathcal{G}_i, \mathcal{G}_j} \mathcal{G}_i \cap \mathcal{G}_j = \emptyset$, and $m_i$ is the mass of token $x_i$. Then, the attention is also refined with $m$ as
\begin{align}
A^\prime_{ij} = \text{softmax}\left(A_{ij} + \log m_j\right),
\label{eq:7}
\end{align}
where $A_{ij}$ is the element in $i$-th row and $j$-th column.
For the mass $m$ of the tokens, ToMe uses the number of constituent tokens. 
Although it has been proven effective in alleviating redundancies, it cannot adaptively adjust the influence of merged tokens considering its importance.
We here propose a \textbf{saliency-aware token merging} that estimates the mass reflecting the saliency of the tokens and then merges the tokens with their corresponding mass to minimize the hindrance induced by the uninformative tokens.
First, we introduce the \textit{background drop} mask to update the mass of foreground and background tokens selectively.
Given the saliency scores, we define the mask as
\begin{align}
M_i = \mathbf{1}_{\{s_i > \bar{s}\}},
\label{eq:8}
\end{align}
where $\mathbf{1}$ is the indicator function and $\bar{s} = \frac{1}{N} \sum^N_i s_i$.
Using the mask above, our framework sets the saliency scores to 0 if a token has a saliency score lower than the average saliency score.
With $M$, we define \textit{masked saliency scores} $\hat{s}$ to focus more on informative tokens among the foreground objects by masking and rescaling the saliency scores as 
\begin{align}
\hat{s}_i = 
\frac{M_i (s_i-\bar{s})}{\max(M_i (s_i-\bar{s}))}.
\label{eq:9}
\end{align}
Yet, we have one more issue while adopting $\hat{s}$ as the mass.
If all tokens in the group $\mathcal{G}_i$ have the saliency scores lower than $\bar{s}$, 
feature aggregation may result in a zero vector losing the entire information.
So, to prevent this, we compute the mass with $\hat{s}$ as 
\begin{align}
\tilde{m}_i = 
\begin{cases}
\hat{s}_i m_{i}, &\text{if } x_i \in X^\text{src}\\
m_{i} &\text{if } x_i \in X^\text{dst}
\end{cases}.
\label{eq:10}
\end{align}
Then, feature merging of \Cref{eq:6} is modified as 
\begin{align}
x^\prime_i = {\sum}_{j\in{\mathcal{G}_i}} \frac{\tilde{m}_j x_j}
{{\sum}_{j^\prime\in{\mathcal{G}_i}} \tilde{m}_{j^\prime}} \text{ and } m^\prime_i = {\sum}_{j\in{\mathcal{G}_i}} \tilde{m}_j,
\label{eq:11}
\end{align}
where $m^\prime_i$ is the mass of $i$-th fused token.
It is worth noting that this saliency-aware token merging can be viewed as feature aggregation only with foreground tokens since the scores of the background tokens are set to 0. In other words, it is a combination of token merging and token pruning. 
Although the same number of tokens are merged in each video, we dynamically adjust the influence of uninformative tokens by suppressing the mess, leading to promising improvements by the proposed saliency-aware token merging, see~\Cref{sec:4.2}.

%% file: Figures/Fig_main.tex
\begin{figure*}[t!]
     \centering
     \includegraphics[trim=0 0 0 0,clip, width=0.9\linewidth]{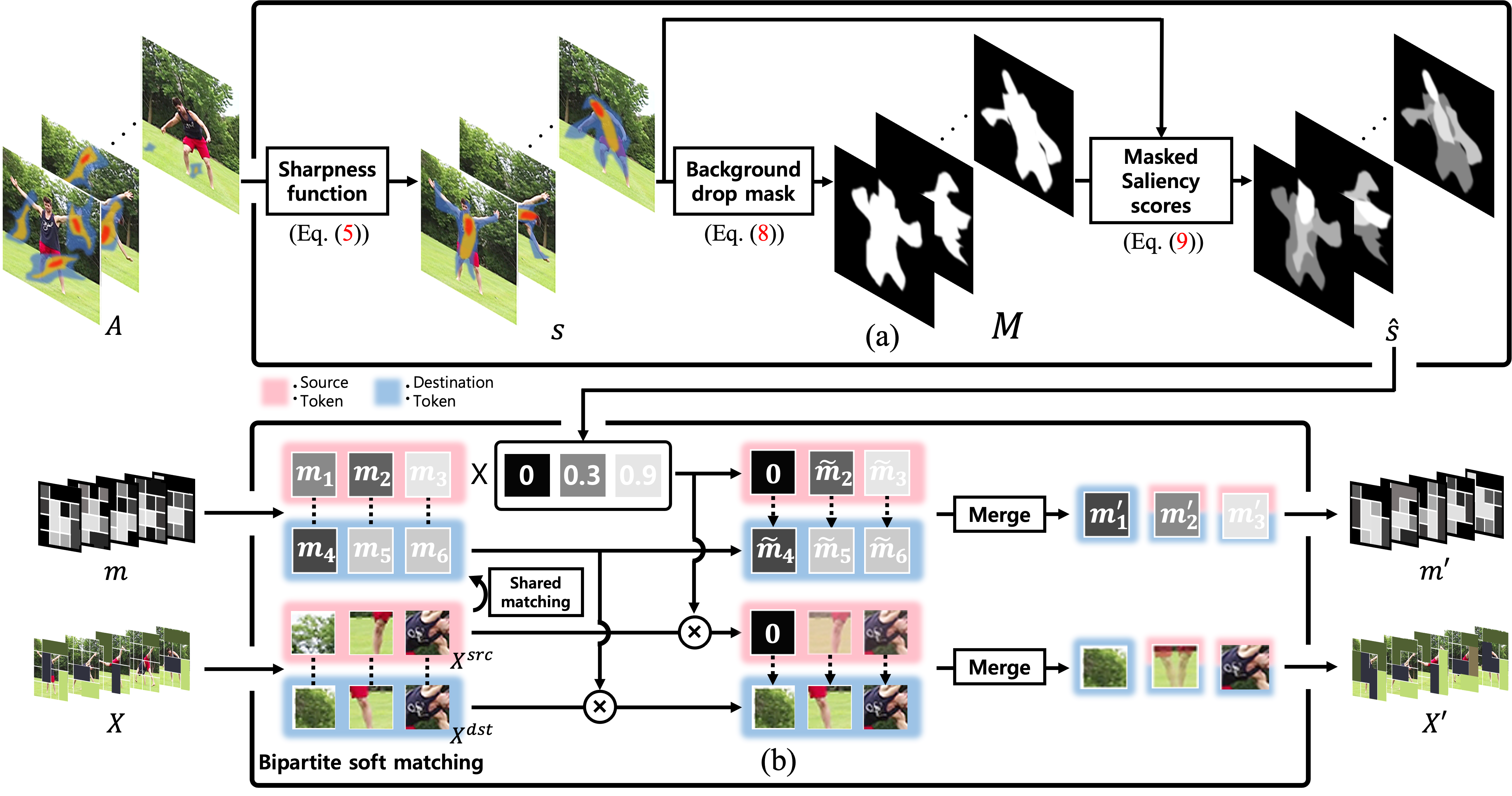}
     \vspace{-5pt}
     \caption{Pipeline of $\textbf{vid-TLDR}$. (a) Given the attention map $A$, the saliency score $s$ is approximated by the $\textit{sharpness function}$ $S$~(\cref{eq:5}). After that, we generate $\textit{background drop}$ mask $M$~(\cref{eq:8}) to minimize the disturbance of background tokens. With $s$ and $M$, we generate $\textit{masked saliency scores}$ $\hat s$~(\cref{eq:9}). (b) Given the tokens $X$ and their corresponding mass $m$, we conduct the matching to group the input tokens. Following that, we update the mass $m$ to $\tilde m$ with $\hat s$~(\cref{eq:10}) to highlight important foreground tokens and minimize the hindrance of background tokens. With updated mass $\tilde{m}$, the grouped tokens are merged into a token $m'$ and $X'$~(\cref{eq:11})}
     \label{fig:3}
\vspace{-15pt}
\end{figure*}

%% file: Figures/Fig_sharpness.tex
\begin{figure}[t!]
     \centering
     \centering
     \includegraphics[trim=0 20 20 0,clip, width=0.9\linewidth]{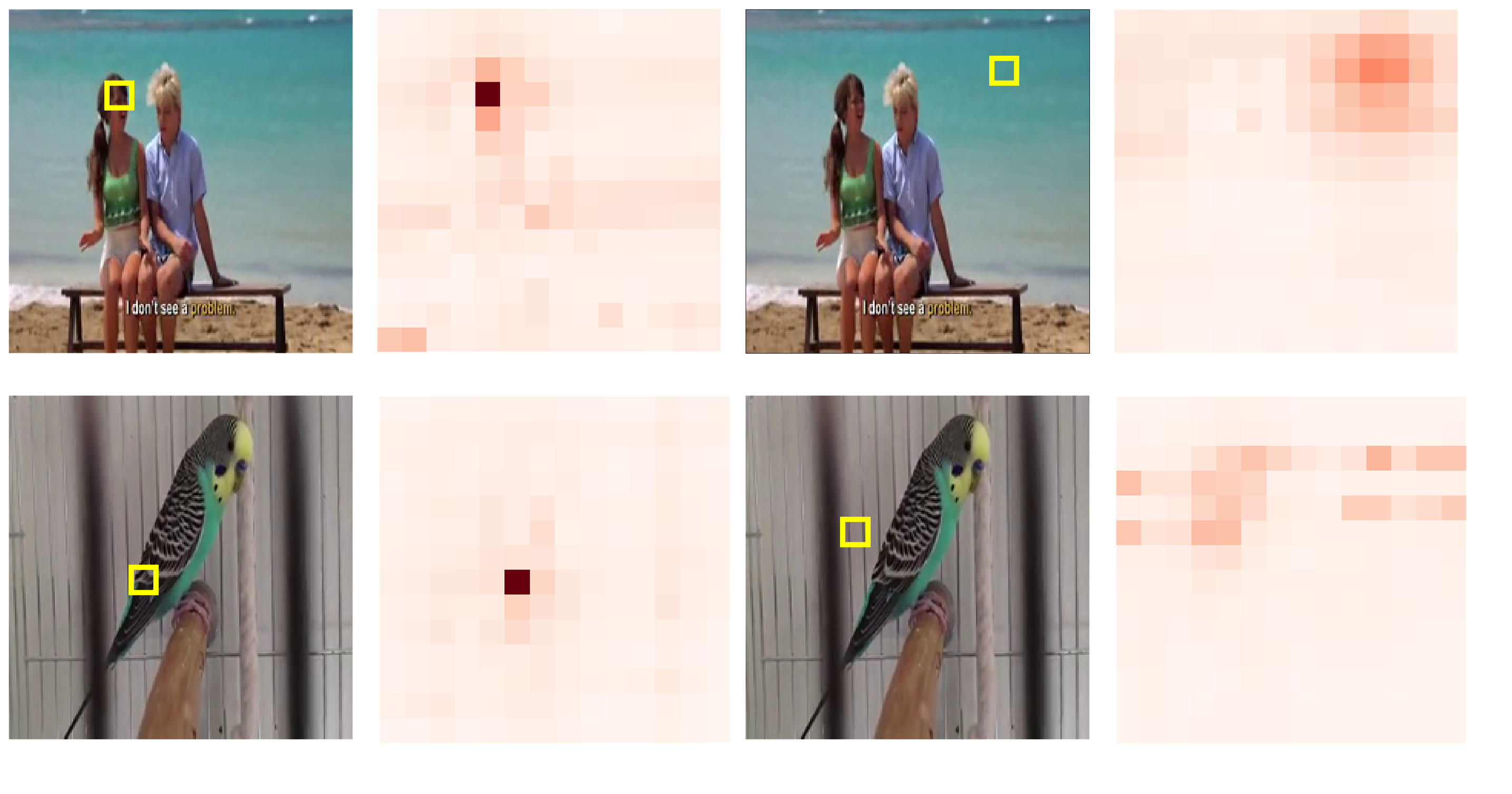}
     \caption{Visualization of attention scores in the first layer of UMT-B~\cite{li2023unmasked}. Given the query $q_i$ denoted as a yellow square, we visualize the attention score $a_i$. Tokens in the foreground objects show a sharper attention map compared to the background tokens.}
     \label{fig:sharp}
\vspace{-15pt}
\end{figure}

%% file: Tables/merging.tex
\begin{table}[t]
\centering
\small
\begin{tabular}{c|c|ccc}
\thickhline
\multicolumn{1}{l|}{\textbf{Layer}} & \multicolumn{1}{c|}{\textbf{GFLOPs}} & \textbf{T2V} & \textbf{V2T} & \textbf{Mean} \\ 
\hline
-  & 303.3  & \textbf{51.0}   & 49.0    & 50.0    \\
\hline
1      & \textbf{237.6}  & 49.6   & \textbf{51.1}    & \textbf{50.4}    \\
2      & 243.1  & 49.6   & 50.8    & 50.2    \\
3      & 248.6  & 49.9   & 50.5    & 50.2    \\
4      & 254.0  & 49.6   & 50.7    & 50.2    \\
5      & 259.5  & 49.7   & 50.6    & 50.2    \\
6      & 265.0  & 49.6   & 50.7    & 50.2    \\
7      & 270.5  & 48.8   & 50.1    & 49.5    \\
\thickhline
\end{tabular}
\caption{
    The comparative study of where to reduce the tokens. 
}
\label{table:merging}
\vspace{-10pt}
\end{table}

%% file: Tables/retrieval.tex
\begin{table*}[t]
\small
\centering
\begin{tabular}{c|l|c|cc|c|cc}
\thickhline
                             \multicolumn{1}{c|}{\multirow{2}{*}{\textbf{Dataset}}} & \multicolumn{1}{c|}{\multirow{2}{*}{\textbf{Metric}}} & \multicolumn{3}{c|}{\textbf{UMT-B}}                                             & \multicolumn{3}{c}{\textbf{UMT-L}}                                             \\
                                & \multicolumn{1}{c|}{} & \multicolumn{1}{c}{Base} & \multicolumn{1}{c}{ToMe} & \multicolumn{1}{c|}{Ours} & \multicolumn{1}{c}{Base} & \multicolumn{1}{c}{ToMe} & \multicolumn{1}{c}{Ours} \\ \hline
\multirow{4}{*}{MSRVTT}         & GFLOPs $\downarrow$   & 303.3  & 231.4                                      & \textbf{178.0}                            & 984.6 & \textbf{529.7}             & 563.1                                   \\ \cline{2-8} 
                                & R@1 $\uparrow$        & 50.0   & 47.0 \scriptsize{($-$3.0})                 & \textbf{50.8}  \scriptsize{($+$0.8)}      & 58.7  & 55.8 \scriptsize{($-$2.9)} & \textbf{58.5} \scriptsize{($-$0.2)}     \\
                                & R@5 $\uparrow$        & 76.8   & 73.2 \scriptsize{($-$3.6})                 & \textbf{75.7}  \scriptsize{($-$1.1)}      & 81.3  & 79.6 \scriptsize{($-$1.7)} & \textbf{81.3} \scriptsize{({$\pm$0.0})} \\
                                & R@10 $\uparrow$       & 83.9   & 82.1 \scriptsize{($-$1.8})                 & \textbf{83.8}  \scriptsize{($-$0.1)}      & 86.8  & 86.1 \scriptsize{($-$0.7)} & \textbf{86.9} \scriptsize{({$+$0.1})} \\ \hline
\multirow{4}{*}{MSVD}           & GFLOPs $\downarrow$   & 303.3  & 218.7                                      & \textbf{181.3}                            & 984.6 & 574.5                      & \textbf{563.1}                          \\ \cline{2-8} 
                                & R@1 $\uparrow$        & \underline{62.1}   & 59.6 \scriptsize{($-$2.5})                 & \textbf{62.7}  \scriptsize{($+$0.6)}      & \underline{70.3}  & 69.5 \scriptsize{($-$0.8)} & \textbf{70.4} \scriptsize{($+$0.1)}     \\
                                & R@5 $\uparrow$        & \underline{84.7}   & 83.8 \scriptsize{($-$0.9})                 & \textbf{84.8}  \scriptsize{($+$0.1)}      & \underline{89.3}  & 88.7 \scriptsize{($-$0.6)} & \textbf{90.5} \scriptsize{($+$1.2)}     \\
                                & R@10 $\uparrow$       & \underline{90.0}   & 89.0 \scriptsize{($-$1.0})                 & \textbf{89.8}  \scriptsize{($-$0.2)}      & \underline{93.2}  & 92.7 \scriptsize{($-$0.5)} & \textbf{94.0} \scriptsize{($+$0.8)}     \\ \hline
\multirow{4}{*}{ActivityNet}    & GFLOPs $\downarrow$   & 303.3  & 236.8                                      & \textbf{227.6}                            & 984.6 & 574.5                      & \textbf{572.9}                          \\ \cline{2-8} 
                                & R@1 $\uparrow$        & 57.2   & 51.7 \scriptsize{($-$5.5})                 & \textbf{56.6}  \scriptsize{($-$0.6)}      & 65.6  & 62.5 \scriptsize{($-$3.1)} & \textbf{65.2} \scriptsize{($-$0.4)}     \\
                                & R@5 $\uparrow$        & 83.7   & 80.7 \scriptsize{($-$3.0})                 & \textbf{83.4}  \scriptsize{($-$0.3)}      & 89.1  & 86.9 \scriptsize{($-$2.2)} & \textbf{88.7} \scriptsize{($-$0.4)}     \\
                                & R@10 $\uparrow$       & 91.6   & 89.6 \scriptsize{($-$2.0})                 & \textbf{91.3}  \scriptsize{($-$0.3)}      & 94.9  & 93.6 \scriptsize{($-$1.3)} & \textbf{94.5} \scriptsize{($-$0.4)}     \\ \hline
\multirow{4}{*}{DiDeMo}         & GFLOPs $\downarrow$   & 303.3  & 241.4                                      & \textbf{212.8}                            & 984.6 & 574.5                      & \textbf{559.0}                          \\ \cline{2-8} 
                                & R@1 $\uparrow$        & \underline{62.1}   & 57.3 \scriptsize{($-$4.8})                 & \textbf{62.4}  \scriptsize{($+$0.3)}      & \underline{70.8}  & 68.0 \scriptsize{($-$2.8)} & \textbf{70.4} \scriptsize{($-$0.4)}     \\
                                & R@5 $\uparrow$        & \underline{86.8}   & 82.6 \scriptsize{($-$4.2})                 & \textbf{86.2}  \scriptsize{($-$0.6)}      & \underline{90.6}  & 89.4 \scriptsize{($-$1.2)} & \textbf{90.5} \scriptsize{($-$0.1)}     \\
                                & R@10 $\uparrow$       & \underline{92.1}   & 89.3 \scriptsize{($-$2.8})                 & \textbf{91.6}  \scriptsize{($-$0.5)}      & \underline{94.5}  & 93.8 \scriptsize{($-$0.7)} & \textbf{94.0} \scriptsize{($-$0.5)}     \\ \hline
\multirow{4}{*}{LSMDC}          & GFLOPs $\downarrow$   & 303.3  & 223.2                                      & \textbf{206.2}                            & 984.6 & 574.5                      & \textbf{583.7}                          \\ \cline{2-8} 
                                & R@1 $\uparrow$        & 32.7   & 27.3 \scriptsize{($-$5.4})                 & \textbf{32.4}  \scriptsize{($-$0.3)}      & 42.2  & 39.2 \scriptsize{($-$3.0)} & \textbf{41.9} \scriptsize{($-$0.3)}     \\
                                & R@5 $\uparrow$        & 54.1   & 49.1 \scriptsize{($-$5.0})                 & \textbf{53.3}  \scriptsize{($-$0.8)}      & 64.9  & 61.6 \scriptsize{($-$3.3)} & \textbf{64.1} \scriptsize{($-$0.8)}     \\
                                & R@10 $\uparrow$       & 63.3   & 57.3 \scriptsize{($-$6.0})                 & \textbf{63.2}  \scriptsize{($-$0.1)}      & 72.3  & 68.7 \scriptsize{($-$3.6)} & \textbf{70.8} \scriptsize{($-$1.5)}     \\ \hline
\multirow{4}{*}{SSV2-label}     & GFLOPs $\downarrow$   & 303.3  & 232.2                                      & \textbf{212.9}                            & 984.6 & 627.2                      & \textbf{610.9}                          \\ \cline{2-8} 
                                & R@1 $\uparrow$        & 64.0   & 60.2 \scriptsize{($-$3.8})                 & \textbf{63.8}  \scriptsize{($-$0.2)}      & 72.4  & 69.9 \scriptsize{($-$2.5)} & \textbf{72.1} \scriptsize{($-$0.3)}     \\
                                & R@5 $\uparrow$        & 88.3   & 86.1 \scriptsize{($-$2.2})                 & \textbf{87.7}  \scriptsize{($-$0.6)}      & 93.4  & 92.2 \scriptsize{($-$1.2)} & \textbf{93.0} \scriptsize{($-$0.4)}     \\
                                & R@10 $\uparrow$       & 92.9   & 91.6 \scriptsize{($-$1.3})                 & \textbf{92.7}  \scriptsize{($-$0.2)}      & 96.7  & 95.8 \scriptsize{($-$0.9)} & \textbf{96.5} \scriptsize{($-$0.2)}     \\ \hline
\multirow{4}{*}{SSV2-Template}  & GFLOPs $\downarrow$   & 303.3  & 241.4                                      & \textbf{203.7}                            & 984.6 & 627.2                      & \textbf{572.9}                          \\ \cline{2-8} 
                                & R@1 $\uparrow$        & 74.6   & 71.8 \scriptsize{($-$2.8})                 & \textbf{74.0}  \scriptsize{($-$0.6)}      & 78.4  & 77.0 \scriptsize{($-$1.4)} & \textbf{78.1} \scriptsize{($-$0.3)}     \\
                                & R@5 $\uparrow$        & 93.9   & \textbf{93.9} \scriptsize{({$\pm$0.0})}    & 93.4           \scriptsize{($-$0.5)}      & 95.9  & 95.1 \scriptsize{($-$0.8)} & \textbf{95.8} \scriptsize{($-$0.1)}     \\
                                & R@10 $\uparrow$       & 96.8   & \textbf{96.6} \scriptsize{($-$0.2)}        & 96.3           \scriptsize{($-$0.5)}      & 97.8  & 97.7 \scriptsize{($-$0.1)} & \textbf{97.9} \scriptsize{($+$0.1)}     \\ \hline
\thickhline
\end{tabular}
\caption{
     Video-text retrieval on MSRVTT~\cite{xu2016msr}, MSVD~\cite{chen-dolan-2011-collecting}, ActivityNet~\cite{caba2015activitynet}, DiDeMo~\cite{anne2017localizing}, LSMDC~\cite{rohrbach2017movie}, SSV2-Label/Template~\cite{lei2022revealing}. Underlined results indicate the number reported with the official repository of UMT~\cite{li2023unmasked} that corrects the misconfiguration of it.
}
\label{table:retrieval}
\vspace{-10pt}
\end{table*}

%% file: section/experiments.tex
\input{Tables/retrieval2}

\section{Experiments}
\label{sec:4}
\noindent\textbf{Baselines.}
To show that vid-TLDR effectively boosts video Transformer, we opt for UMT~\cite{li2023unmasked} as the baseline, which achieves state-of-the-art performance on various video tasks.
Since vid-TLDR is the training-free plug-in module, we simply apply it right after the self-attention in the early layer of UMT and evaluate it without any additional training. 
We conduct vid-TLDR in the first four layers. 
The detailed reduced number of tokens for each dataset is provided in the supplement.
In multi-modal tasks, we adopt vid-TLDR on the vision encoder of UMT.
Except for the reduced number of tokens, we conduct whole experiments under the same evaluation settings of UMT.
We also report the results with the previous merging method, ToMe~\cite{bolya2022token}, which can be added to the pre-trained video Transformer. 
For the settings of ToMe, we respect the default setups, where the same number of tokens are merged based on the similarity in every layer.
For a fair comparison, we try to maintain similar FLOPs.


\input{Tables/qa}
\subsection{Experimental results.}
\noindent\textbf{Video-text retrieval}
First, we summarize the results of video-text retrieval with MSRVTT~\cite{xu2016msr}, MSVD~\cite{chen-dolan-2011-collecting}, ActivityNet~\cite{caba2015activitynet}, and DiDeMo~\cite{anne2017localizing}, LSMDC~\cite{rohrbach2017movie}, Something-Something~\cite{lei2022revealing}.
Video-text retrieval contains two subtasks: video-to-text retrieval, and text-to-video retrieval.
Video-to-text retrieval is to find the most relevant text concerning the given video, while text-to-video retrieval is conducted in the opposite direction. 
We report the average of the results in~\Cref{table:retrieval}.
As summarized, our proposed method consistently shows competitive performances compared to base UMT~\cite{li2023unmasked} and outperforms ToMe\cite{bolya2023tomesd} in every backbone and dataset.
Compared to ToMe, vid-TLDR shows a performance gap of (+4.0\%, +2.1\%) on average R@1 in UMT-B, UMT-L even with the lower FLOPs.
Further, vid-TLDR with UMT-B even surpasses the base model with the improvements of (+0.8\%, +0.6\%, +0.3\%) R@1 while reducing FLOPs by (41.3\%, 40.2\%, 29.8\%) in MSRVTT, MSVD, DiDeMo, respectively. 
And, we also observe that vid-TLDR achieves competitive performance with base UMT-L despite the much lower FLOPs.
We further provide the comparison with other video backbones in text-to-video retrieval (\Cref{table:retrieval2}).
For reporting the table, we experiment with the model used in~\Cref{table:retrieval}.
Although we largely reduce the computational cost of UMT-B, and UMT-L by 34.1\%, 42.7\% on average, they still show superior performance compared to other video backbones in MSRVTT, DiDeMo, and ActivityNet.

\noindent\textbf{Video question answering.}
We experiment with video question answering with MSR-QA~\cite{DBLP:conf/mm/XuZX0Z0Z17} and MSVD-QA~\cite{DBLP:conf/mm/XuZX0Z0Z17}, summarizing the results in~\Cref{table:qa}. In MSR-QA, the results of UMT-B and UMT-L are 44.9\% and 47.1\%, respectively. After adopting vid-TLDR on each model, we achieved the competitive performance of 44.8\% and 47.0\% while reducing FLOPs by 37.9\% in UMT-B, and  42.1\% in UMT-L. Similarly, we could lessen the computational cost in MSVD-QA with the small performance degradation despite the much lower FLOPs. 
Specifically, the performance drop is only 0.1\% and 0.3\% in UMT-B and UMT-L with much lower FLOPs compared to the original model.
To summarize, vid-TLDR boosts the model with a minor accuracy drop in video question answering as well as video-text retrieval.

\input{Tables/videomae}
\input{Tables/vifi}
\noindent\textbf{vid-TLDR with other video Transformers and tasks.}
To demonstrate the generalizability of vid-TLDR, we apply vid-TLDR to other video Transformers: VideoMAE~\cite{tong2022videomae}, and ViFi-CLIP~\cite{rasheed2023fine}. 
The results of action recognition with VideoMAE are presented in~\Cref{tab:videomae}.
We use Something Something V2 (\textbf{SSV2})~\cite{goyal2017something} and UCF101~\cite{soomro2012ucf101}.
Experimental results are promising that vid-TLDR lessens the complexity of VideoMAE by almost 70\% (180.5 (G) $\rightarrow$ 56.1 (G)) with a minimal performance degradation compared to ToMe (\eg, 14.4\% (75.7 $\rightarrow$ 90.1) improvement over ToMe). 
Further, we adopt vid-TLDR on ViFi-CLIP in the base-to-novel generalization tasks (\ie, training only with the base classes, then evaluating on both seen (base). 
As shown in \Cref{tab:vifi} with UCF-101, vid-TLDR halving the FLOPs (563 $\rightarrow$ 279) of ViFi-CLIP surpassing ToMe with a 19.4\% (55.3\% $\rightarrow$ 74.7\%) gain in the harmonic mean (\textbf{HM}).

\subsection{Ablation studies}
\label{sec:4.2}
In this section, we provide the ablations studies of vid-TLDR. 
We studied the effectiveness of each component with UMT-B~\cite{li2023unmasked} and MSRVTT~\cite{xu2016msr} (\Cref{table:ablation}). 
The first row of the table indicates the base model without any token reduction.
First, regarding the metric for informativeness, we have compared our proposed saliency scores $s$ in~\Cref{eq:9} with the attentiveness $\bar{a}$ in~\Cref{eq:3}, and attention rollout $\tilde{a}$ in~\Cref{eq:4}.
For comparison, based on each metric, we simply prune the token having the low scores, equal to the number used for reporting~\Cref{table:retrieval}.
As shown in the table, since the tokens are dropped in the earlier layers, the attentiveness and attention rollout may drop the salient tokens resulting in a substantial performance drop in both text-to-video retrieval and video-to-text retrieval.
Specifically, the performance degradation on average is 1.0\%, 0.7\% when using $\bar{a}$ and $\tilde{a}$.
Further, although attention rollout shows slightly better performance than simple attentiveness, it shows worse FLOPS due to the repeated forward process for estimating attention flows.
On the other hand, by dropping the tokens based on our saliency scores $s$ estimated by the sharpness of attention, we could lessen the FLOPs with superior performance compared to base.
Finally, introducing the saliency-aware token merging, we have achieved the +0.8\% (50.0\% $\rightarrow$ 50.8\%) gain despite the 58.7\% FLOPs compared to the original UMT.

\input{Tables/ablation}

\input{Figures/Fig_temporal}
\input{Figures/QA_Fig}

\subsection{Analysis}
\noindent\textbf{Temporal bias.} 
As we discussed in in~\Cref{sec:2}, the video Transformers contains the temporal bias that neglects the later frames of a video.
For a better understanding, we here quantitatively compare three metrics, attentiveness $\bar{a}$ (Att.), attention rollout $\tilde{a}$ (ATT. Roll.), and our saliency scores $\hat{s}$. 
We measure each score with UMT-B~\cite{li2023unmasked} in MSRVTT~\cite{xu2016msr}.
After normalizing the score across all frames, we calculate the sum of the scores for each frame.
In other words, it represents the ratio of scores per each frame concerning the total score. (see~\Cref{fig:temporal})
In both attentiveness and attention rollout, the earlier frames show higher scores compared to later frames, specifically, the score of the first frame is almost $\times 3$ times higher than the last frame.
As a result, if we rely on these metrics, it is prone to merge the tokens in the later frame without considering their informativeness of them.
Whereas, as shown in the figure, our proposed $\hat{s}$ shows the more robust ratio across the frame, capturing the informative tokens even in the last frame.

\noindent\textbf{Visualization of vid-TLDR results.}
In~\Cref{fig:qa}, we show the qualitative results to understand the behavior of vid-TLDR. 
Given video clips of MSRVTT, we visualize the merged token. We further provide the heatmaps for analyzing the mass of each token. More precisely, we divide the mass of merged tokens by the number of constituent tokens to represent the mass concerning each input token.
As shown in the figure, through the saliency-aware token merging, the foreground object shows a much higher mass than the background tokens.
In short, through~\Cref{eq:7}, our vid-TLDR minimizes the hindrance from the background tokens during the self-attention layer.

%% file: Tables/retrieval2.tex

\begin{table}[t!]
\centering
\small
\setlength{\tabcolsep}{3pt}
\begin{tabular}{l|r|cccc}
\thickhline
\multicolumn{1}{c|}{\textbf{Method}} & \multicolumn{1}{c|}{\textbf{\#Pairs}} & \textbf{MSR.} & \textbf{MSVD} & \textbf{Act.} & \textbf{DiDe.} \\
\hline 
ClipBERT~\cite{lei2021less}             & 5.4M  & 22.0           & -                & 21.3          & 20.4                      \\
Frozen~\cite{Bain21}                    & 5M    & 31.0           & 33.7             & -             & 34.6                      \\
VIOLET~\cite{fu2021violet}              & 138M  & 34.5           & -                & -             & 32.6                      \\
All-in-one~\cite{wang2023all}           & 138M  & 37.9           & -                & 22.4          & 32.7                      \\
LAVENDER~\cite{li2023lavender}          & 30M   & 40.7           & 50.1             & -             & 53.4                      \\
Singularity~\cite{lei2022revealing}     & 17M   & 42.7           & -                & 48.9          & 53.1                      \\
OmniVL~\cite{wang2022omnivl}            & 17M   & 47.8           & -                &  -            & 52.4                      \\
VINDLU~\cite{cheng2023vindlu}           & 25M   & 46.5           & -                & 55.0          & 61.2                      \\
CLIP4Clip~\cite{luo2022clip4clip}       & 400M  & 44.5           & 46.2             & 40.5          & 42.8                      \\
CLIP-ViP~\cite{xue2022clip}             & 500M  & 54.2           & -                & 53.4          & 50.5                      \\
InternVideo~\cite{wang2022internvideo}  & 646M  & 55.2           & \textbf{58.4}    & 62.2          & 57.9                      \\ \hline
UMT-B~\cite{li2023unmasked}             & 25M   & 51.0           & \underline{50.8} & 58.3          & \underline{63.7}          \\
UMT-L~\cite{li2023unmasked}             & 25M   & \textbf{58.8}  & \underline{58.2} & \textbf{66.8} & \underline{\textbf{72.5}} \\ \hline
UMT-B-Ours                              & 25M   & 50.9           & 50.5             & 57.8          & 64.1             \\
UMT-L-Ours                              & 25M   & 58.1           & 57.9             & 66.7          & 72.3             \\ \thickhline
\end{tabular}
\caption{
    Text-to-video retrieval on MSRVTT (MSR.)~\cite{xu2016msr}, DiDeMo (DiDe.)~\cite{anne2017localizing}, ActivityNet (Act.)~\cite{caba2015activitynet}, MSVD~\cite{chen-dolan-2011-collecting}. “\#Pairs” denotes the number of pre-training pairs. We use the models of each dataset in~\Cref{table:retrieval} to report ours.
}
\vspace{-15pt}
\label{table:retrieval2}
\end{table}


%% file: Tables/qa.tex
\begin{table}[t]
\centering
\small
\setlength{\tabcolsep}{3pt}
\begin{tabular}{l|r|r|cc}
\thickhline
\multicolumn{1}{c|}{\multirow{2}{*}{\textbf{Method}}} & \multicolumn{1}{l|}{\multirow{2}{*}{\textbf{\#Pairs}}} & \multicolumn{1}{l|}{\multirow{2}{*}{\textbf{GFLOPs}}} & \multirow{2}{*}{\textbf{MSR-QA}} & \multirow{2}{*}{\textbf{MSVD-QA}} \\
                                  & \multicolumn{1}{l|}{}                                  & \multicolumn{1}{l|}{}                                 &                                  &                                   \\ \hline
ALPRO~\cite{li2022align}               & 5M     & 392.5           & 42.1            & 45.9           \\
JustAsk~\cite{yang2021just}            & 69M    & 340.7           & 41.5            & 47.5           \\
All-in-one~\cite{wang2023all}          & 138M   & 1017.0          & 44.3            & 47.9           \\
MERLOT~\cite{zellers2021merlot}        & 180M   & -               & 43.1            & -              \\
VIOLET~\cite{fu2021violet}             & 138M   & 282.0           & 43.9            & 47.9           \\
Singularity~\cite{lei2022revealing}    & 17M    & 211.0           & 43.9            & -              \\
OmniVL~\cite{wang2022omnivl}           & 17M    & -               & 44.1            & 51.0           \\
VINDLU~\cite{cheng2023vindlu}          & 25M    & 278.5           & 44.6            & -              \\
FrozenBiLM~\cite{yang2022zero}         & 400M   & 340.7           & 47.0            & 54.8           \\
InternVideo~\cite{wang2022internvideo} & 646M   & 666.2           & \textbf{47.1}   & 55.5           \\
VideoCoCa~\cite{yan2022video}          & 4.8B   & 29820           & 46.0            & \textbf{56.9}  \\ \hline
UMT-B~\cite{li2023unmasked}            & 25M    & 303.3           & 44.9            & 49.5           \\
UMT-L~\cite{li2023unmasked}            & 25M    & 984.6           & \textbf{47.1}   & 55.2           \\ \hline
UMT-B-Ours                             & 25M    & \textbf{188.5}  & 44.8            & 49.4           \\
UMT-L-Ours                             & 25M    & 569.8           & 47.0            & 54.9           \\
\thickhline
\end{tabular}
\caption{
   Video question-answering on MSRVTT-QA~\cite{DBLP:conf/mm/XuZX0Z0Z17} \& MSVD-QA~\cite{DBLP:conf/mm/XuZX0Z0Z17}.
}
\label{table:qa}
\vspace{-10pt}
\end{table}

%% file: Tables/videomae.tex
\begin{table}[t!]
  \centering 
  \small
    \centering
    \begin{tabular}{l|cc|cc}
    \thickhline
         \multicolumn{1}{c|}{\multirow{2}{*}{\textbf{Method}}}&  \multicolumn{2}{c|}{\textbf{UCF101}} & \multicolumn{2}{c}{\textbf{SSV2}}  \\
         & \multicolumn{1}{c}{GFLOPs} & \multicolumn{1}{c|}{Acc} & \multicolumn{1}{c}{GFLOPs} & \multicolumn{1}{c}{Acc}\\
        \midrule
        VideoMAE~\cite{tong2022videomae} & 180.5  & 91.3 & 180.5 & 70.8 \\
        +ToMe~\cite{bolya2022token}& 58.4 & 75.7 & 58.4  & 58.5  \\
        +vid-TLDR & 56.1 & 90.1  & 56.6   & 69.6 \\
    \thickhline
  \end{tabular}
  \vspace{-5pt}
  \caption{Action recognition with VideoMAE on UCF101~\cite{soomro2012ucf101} \& Something Something V2~\cite{goyal2017something}}
  \label{tab:videomae}
\vspace{-5pt}
\end{table}

%% file: Tables/vifi.tex
\begin{table}[t!]

  \centering 
    \centering
    \small
    \begin{tabular}{l|cccc}
    \thickhline
         \textbf{Method} & \multicolumn{1}{c}{\textbf{FLOPs}} & \multicolumn{1}{c}{\textbf{Base}} & \multicolumn{1}{c}{\textbf{Novel}} & \multicolumn{1}{c}{\textbf{HM}}\\
        \midrule
        ViFi-CLIP~\cite{rasheed2023fine} & 563  & 92.9 & 67.7 & 78.3 \\
        +ToMe~\cite{bolya2022token}& 279 & 72.0 & 44.8 & 55.3 \\
        +vid-TLDR & 279  & 91.3 & 63.2  & 74.7 \\
    \thickhline
  \end{tabular}
  \vspace{-5pt}
  \caption{Base-to-novel generalization on UCF101~\cite{soomro2012ucf101}}
  \label{tab:vifi}
\vspace{-10pt}
\end{table}

%% file: Tables/ablation.tex
\begin{table}[t]
\centering
\setlength{\tabcolsep}{2pt}
{
\small
\begin{tabular}{ccc|c|cccc}
\thickhline
$\bar{a}$ & $\tilde{a}$ & $s$ & S.A. ToMe & GFLOPS & T2V & V2T & Mean\\
\hline
-&-&-&-&303.3 & \textbf{51.0} & 49.0 & 50.0\\
\hline
\checkmark &            &            &                        & \textbf{175.9} & 50.8  & 47.2  & 49.0 \\
           & \checkmark &            &                        & 479.2 & 50.2  & 48.3  & 49.3 \\
           &            & \checkmark &                        & \textbf{175.9} & 50.6  & 50.0  & 50.3 \\ \hline
           &            & \checkmark & \checkmark & 178   & 50.9  & \textbf{50.7}  & \textbf{50.8} \\

\thickhline
\end{tabular}}
\caption{
    Ablations studies on vid-TLDR. The first row indicates the base UMT-B~\cite{li2023unmasked}. S.A. ToMe denotes the saliency-aware token merging.
}
\vspace{-10pt}
\label{table:ablation}
\end{table}

%% file: Figures/Fig_temporal.tex
\begin{figure}[t!]
     \centering
     \includegraphics[trim=0 0 0 0,clip, width=0.80\linewidth]{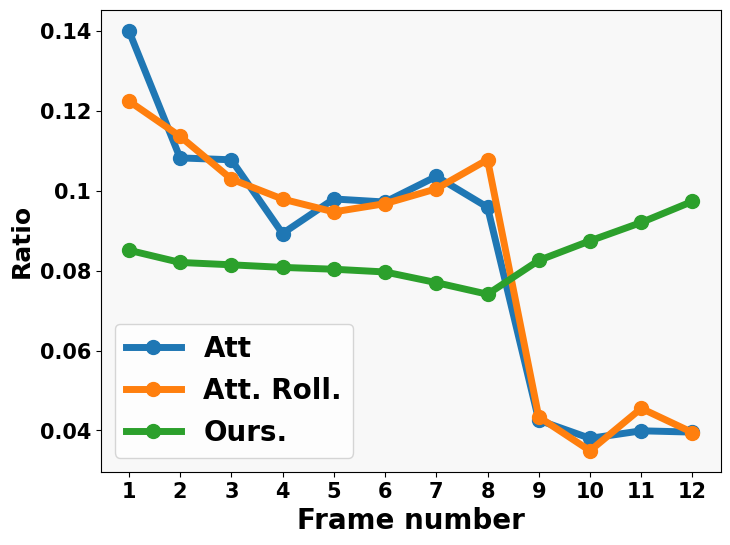}
     \caption{The ratio of the sum of the scores in each frame. Given the informativeness scores in the first layer of the video Transformer, we extract the sum of scores in each frame and then calculate the ratio to the total scores across all the frames.}
     \label{fig:temporal}
\vspace{-10pt}
\end{figure}

%% file: Figures/QA_Fig.tex
\begin{figure*}[t!]
     \centering
     \begin{subfigure}[b]{\linewidth}
         \centering
         \includegraphics[trim=0 0 0 0,clip, width=\linewidth]{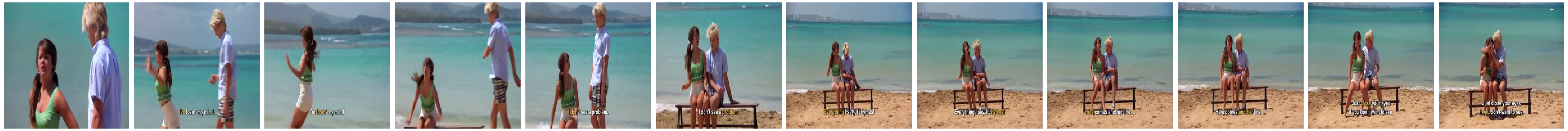}
     \end{subfigure}
     \hfill
     \begin{subfigure}[b]{\linewidth}
         \centering
         \includegraphics[trim=0 0 0 0,clip, width=\linewidth]{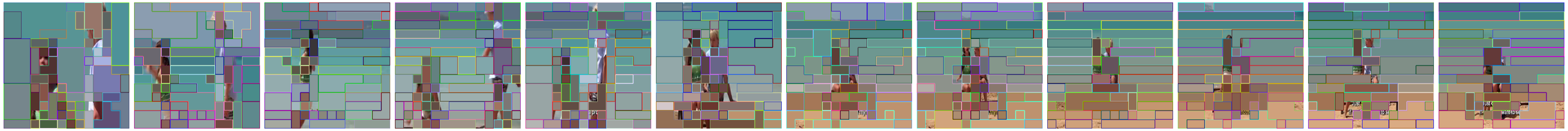}
     \end{subfigure}
     \hfill
     \begin{subfigure}[b]{\linewidth}
         \centering
         \includegraphics[trim=0 0 0 0,clip, width=\linewidth]{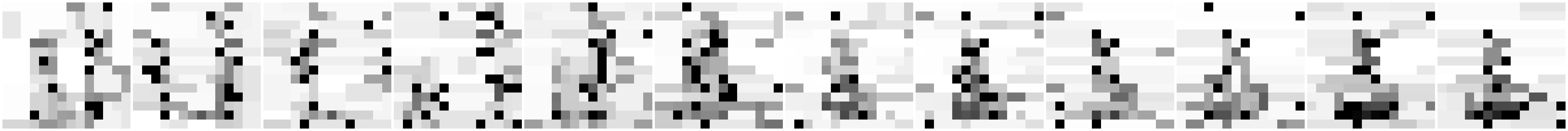}
     \end{subfigure}
     \caption{Qualitative results of vid-TLDR. Given video clips, we visualize the merged tokens and their corresponding mass.
     }
     \label{fig:qa}
\vspace{-15pt}
\end{figure*}

%% file: section/relatedworks.tex
\section{Related works}
\noindent\textbf{Vision Transformer.}
ViT~\cite{dosovitskiy2020image} has become one of the most popular and basic components in computer vision along with CNNs. With the surge of large-scale datasets, its lower inductive bias compared to CNNs endows it with robust generalization, leading to another successive adoption on several downstream tasks in computer vision, including classification~\cite{dosovitskiy2020image,touvron2021training}, detection~\cite{carion2020end,wang2022anchor,meng2021-CondDETR,zhu2020deformable,li2022dn,zang2022open}  segmentation~\cite{wang2020end,xie2021segformer}, image encoding~\cite{he2022masked,bachmann2022multimae,wudenoising} for generation~\cite{lee2021vitgan}.
Although Transformer shows promising results on computer vision the attention mechanism~\cite{bahdanau2014neural} incurs quadratic complexity and restrains its scalability. So, there have been many attempts to mitigate this problem. These attempts can be categorized as ameliorating model architecture itself, leveraging pre-trained models, or reducing the input tokens. For example, \cite{katharopoulos_et_al_2020,vyas_et_al_2020,wang2020linformer,kitaev2020reformer,zaheer2020bigbird} focused on the attention mechanism itself to approximate the complexity to linear.  
More recent works enabled acceleration even without model modification by pruning~\cite{meng2022adavit,kim2022learned,lee2022sparse} or merging~\cite{bolya2022token,bolya2023tomesd,marin2021token,pan2022less} input tokens, with minimal performance degradation. 

\noindent\textbf{Video understanding.}
Video understanding is not the same as the image, since frames of video are not independent images, but highly related to each other in the temporal axis.
Prior works have leveraged transformers for understanding video, including retrieval~\cite{fu2021violet,lei2022revealing,wang2022omnivl,cheng2023vindlu,luo2022clip4clip,yan2022video,li2023unmasked,wang2023all}, question answering~\cite{li2022align,yang2021just,wang2023all,zellers2021merlot,lei2022revealing,cheng2023vindlu,wang2022omnivl,wang2022internvideo,yan2022video}, captioning~\cite{lin2021end-to-end,Gu_2023_CVPR} and representation learning~\cite{tong2022videomae,wang2023videomae,huang2023mgmae}.
Especially, along with the success of image foundation models, some works~\cite{rasheed2023fine, ma2022x} leverage CLIP~\cite{radford2021learning}, which is pre-trained ViTs with the large-scale image-text pairs, for understanding video.
Also, recent studies~\cite{zellers2021merlot,tong2022videomae,wang2023videomae,li2023unmasked,wang2022internvideo} focus on scaling the Transformers for video foundation models to utilize the flexibility for multi-modal tasks. 
Yet, despite the intensive computational cost caused by the massive number of tokens, efficient video Transformers are less explored. 
\label{sec:5}

%% file: section/conclusion.tex
\section{Conclusion}
\label{sec:6}
In this paper, we propose vid-TLDR, training free token merging for light-weight video Transformer. We demonstrate the necessity of performing early token merging in video Transformers and delineate the associated challenges. To address these challenges, we design the new token-merging mechanism as follows. First, we devise a better saliency detector using attention sharpness which can localize salient regions even in the early layers of the Transformer and mitigate the temporal biases of video Transformers. In addition, we revise the mass of token merging so that the influence of uninformative tokens is suppressed and the importance of tokens within the foreground objects is taken into account. The experiments show that our method consistently outperforms previous merging methods in every backbone and dataset even with lower computation, verifying both the efficacy and efficiency of our method in video Transformers. 

\section*{Acknowledgments}
This work was partly supported by ICT Creative Consilience Program through the Institute of Information \& Communications Technology Planning \& Evaluation (IITP) grant funded by the Korea government (MSIT)(IITP-2024-2020-0-01819), 
the National Research Foundation of Korea (NRF) grant funded by the Korea government (MSIT)(NRF-2023R1A2C2005373), and Electronics and Telecommunications Research Institute (ETRI) grant funded by the Korean government (24ZB1200, Fundamental Technology Research for Human-Centric Autonomous Intelligent Systems).

%% file: rebuttal/fig_att.tex
 \begin{figure}[h!]
    \centering
       \includegraphics[trim=0 0 0 0,clip,width=0.95\linewidth]{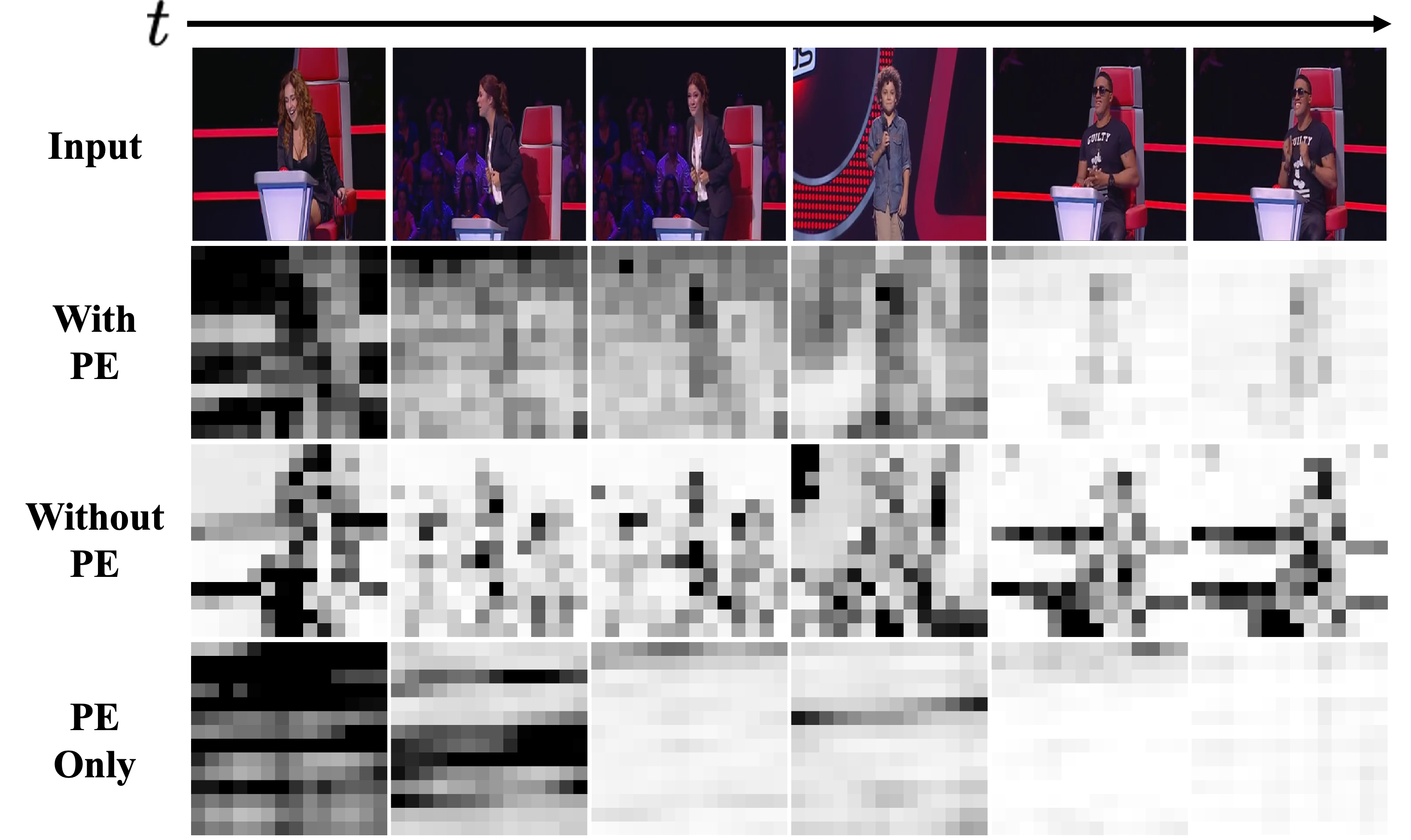}
       \vspace{-10pt}
\caption{Visualization of temporal bias caused by PE.}
\label{fig:att}
\vspace{-10pt}
 \end{figure}

%% file: Supple/reduced_r.tex
\begin{table}[h]
\centering
\caption{
    \textbf{Number of tokens reduced in each experiment.}
}
\centering
\small
\setlength{\tabcolsep}{2pt}
\begin{tabular}{c|c|l|ccccc}
\thickhline
\multirow{2}{*}{Task}      & \multirow{2}{*}{Method}   &  \multirow{2}{*}{Dataset} & \multicolumn{5}{c}{Block index}   \\
                           &                           &                           &   1   &  2    &  3    &  4    &  5 - Top  \\ \hline
\multirow{8}{*}{Retrieval} & \multirow{7}{*}{UMT-B}    &      MSRVTT               &  600  &  200  &  100  &  0    &  0    \\
                           &                           &      MSVD                 &  450  &  250  &  200  &  0    &  0    \\
                           &                           &      ActivityNet          &  500  &  0    &  0    &  0    &  0    \\
                           &                           &      DiDeMo               &  250  &  150  &  150  &  150  &  0    \\
                           &                           &      LSMDC                &  400  &  150  &  150  &  0    &  0    \\
                           &                           &      SSV2-Label           &  200  &  200  &  200  &  100  &  0    \\
                           &                           &      SSV2-Template        &  400  &  300  &  0    &  0    &  0    \\ \cline{2-8} 
                           & \multirow{7}{*}{UMT-L}    &      MSRVTT               &  550  &  400  &  0    &  0    &  0    \\
                           &                           &      MSVD                 &  600  &  200  &  100  &  100  &  0    \\
                           &                           &      ActivityNet          &  600  &  300  &  0    &  0    &  0    \\
                           &                           &      DiDeMo               &  800  &  100  &  0    &  0    &  0    \\
                           &                           &      LSMDC                &  250  &  250  &  200  &  200  &  0    \\
                           &                           &      SSV2-Label           &  450  &  250  &  100  &  0    &  0    \\
                           &                           &      SSV2-Template        &  350  &  300  &  100  &  50   &  0    \\  \hline
\multirow{4}{*}{VQA}       & \multirow{2}{*}{UMT-B}    &      MSRVTT               &  600  &  200  &  0    &  0    &  0    \\
                           &                           &      MSVD                 &  600  &  200  &  0    &  0    &  0    \\ \cline{2-8} 
                           & \multirow{2}{*}{UMT-L}    &      MSRVTT               &  400  &  400  &  100  &  0    &  0    \\
                           &                           &      MSVD                 &  400  &  400  &  100  &  0    &  0    \\
                           
\thickhline
\end{tabular}
\label{sup:reduced_r}
\end{table}

%% file: Tables/retrieval_t2v.tex
\begin{table*}[t]
\centering
\caption{
     \textbf{Detailed results on text-to-video \& video-to-text retrieval.}
}
\label{tab:retrieval_t2v}
\begin{tabular}{c|c|c|c|cc|c|cc}
\thickhline
\multicolumn{1}{c|}{\multirow{2}{*}{\textbf{Dataset}}} & \multicolumn{1}{c|}{\multirow{2}{*}{\textbf{Metric}}} & \multicolumn{1}{c|}{\multirow{2}{*}{\textbf{Type}}} & \multicolumn{3}{c|}{\textbf{UMT-B}} & \multicolumn{3}{c}{\textbf{UMT-L}} \\
 & \multicolumn{1}{c|}{} & &\multicolumn{1}{c}{Base} & \multicolumn{1}{c}{ToMe} & \multicolumn{1}{c|}{Ours} & \multicolumn{1}{c}{Base} & \multicolumn{1}{c}{ToMe} & \multicolumn{1}{c}{Ours} \\ \hline
\multirow{7}{*}{MSRVTT}         & GFLOPs $\downarrow$                   & (G) & 303.3  & 231.4 & 178.0   & 984.6 & 529.7  & 563.1 \\ \cline{2-9} 
                                & \multirow{2}{*}{{R@1 $\uparrow$}}     & T2V & 51.0   & 46.6  & 50.9    & 58.8  & 55.6   & 58.1  \\
                                &                                       & V2T & 49.0   & 47.4  & 50.7    & 58.6  & 56.0   & 58.7  \\ \cline{2-9}
                                & \multirow{2}{*}{{R@5 $\uparrow$ }}    & T2V & 76.5   & 71.8  & 75.8    & 81.0  & 79.7   & 81.0  \\
                                &                                       & V2T & 77.0   & 74.6  & 75.5    & 81.6  & 79.4   & 81.6  \\ \cline{2-9}
                                & \multirow{2}{*}{{R@10 $\uparrow$}}    & T2V & 84.2   & 80.9  & 83.9    & 87.1  & 85.3   & 86.8  \\
                                &                                       & V2T & 84.7   & 83.2  & 83.9    & 86.5  & 86.8   & 86.9  \\ \hline
\multirow{7}{*}{MSVD}           & GFLOPs $\downarrow$                   & (G) & 303.3  & 218.7 & 181.3   & 984.6 & 574.5  & 563.1 \\  \cline{2-9} 
                                & \multirow{2}{*}{{R@1 $\uparrow$}}     & T2V & 50.8   & 47.8  & 50.5    & 58.2  & 57.1   & 57.9  \\
                                &                                       & V2T & 73.3   & 71.3  & 74.9    & 82.4  & 81.9   & 82.7  \\ \cline{2-9}
                                & \multirow{2}{*}{{R@5 $\uparrow$ }}    & T2V & 79.7   & 78.2  & 79.7    & 83.9  & 83.6   & 83.8  \\
                                &                                       & V2T & 88.8   & 89.4  & 90.0    & 94.6  & 93.9   & 94.5  \\ \cline{2-9}
                                & \multirow{2}{*}{{R@10 $\uparrow$}}    & T2V & 85.7   & 85.5  & 86.3    & 89.6  & 89.4   & 89.4  \\
                                &                                       & V2T & 93.7   & 92.5  & 93.3    & 96.7  & 96.0   & 96.3  \\ \hline
\multirow{7}{*}{ActivityNet}    & GFLOPs $\downarrow$                   & (G) & 303.3  & 236.8 & 227.6   & 984.6 & 574.5  & 572.9 \\  \cline{2-9} 
                                & \multirow{2}{*}{{R@1 $\uparrow$}}     & T2V & 58.3   & 51.9  & 57.8    & 66.8  & 63.3   & 66.7  \\
                                &                                       & V2T & 56.0   & 51.4  & 55.4    & 64.4  & 61.7   & 63.9  \\ \cline{2-9}
                                & \multirow{2}{*}{{R@5 $\uparrow$ }}    & T2V & 83.9   & 80.8  & 83.7    & 89.1  & 87.3   & 88.6  \\
                                &                                       & V2T & 83.5   & 80.6  & 83.2    & 89.1  & 86.5   & 88.7  \\ \cline{2-9}
                                & \multirow{2}{*}{{R@10 $\uparrow$}}    & T2V & 91.5   & 89.4  & 91.5    & 94.9  & 93.6   & 94.4  \\
                                &                                       & V2T & 91.7   & 89.9  & 91.1    & 94.8  & 93.6   & 94.5  \\ \hline
\multirow{7}{*}{DiDeMo}         & GFLOPs $\downarrow$                   & (G) & 303.3  & 218.7 & 212.8   & 984.6 & 574.5  & 559.0 \\  \cline{2-9} 
                                & \multirow{2}{*}{{R@1 $\uparrow$}}     & T2V & 63.7   & 56.4  & 64.1    & 72.5  & 68.7   & 72.3  \\
                                &                                       & V2T & 60.5   & 58.2  & 60.8    & 69.2  & 67.3   & 68.5  \\ \cline{2-9}
                                & \multirow{2}{*}{{R@5 $\uparrow$ }}    & T2V & 87.8   & 82.3  & 87.2    & 91.2  & 89.5   & 91.2  \\
                                &                                       & V2T & 85.7   & 82.8  & 85.1    & 89.9  & 89.2   & 89.8  \\ \cline{2-9}
                                & \multirow{2}{*}{{R@10 $\uparrow$}}    & T2V & 93.5   & 89.0  & 92.7    & 94.8  & 94.6   & 94.2  \\
                                &                                       & V2T & 90.7   & 89.5  & 90.4    & 94.1  & 92.9   & 93.8 \\ \hline
\multirow{7}{*}{LSMDC}          & GFLOPs $\downarrow$                   & (G) & 303.3  & 223.2 & 206.2   & 984.6 & 574.5  & 583.7 \\  \cline{2-9} 
                                & \multirow{2}{*}{{R@1 $\uparrow$}}     & T2V & 32.7   & 25.6  & 32.9    & 43.0  & 39.7   & 43.1  \\
                                &                                       & V2T & 32.7   & 29.0  & 31.8    & 41.4  & 38.7   & 40.7  \\ \cline{2-9}
                                & \multirow{2}{*}{{R@5 $\uparrow$ }}    & T2V & 54.7   & 47.6  & 53.6    & 65.5  & 61.8   & 64.5  \\
                                &                                       & V2T & 53.5   & 50.5  & 52.9    & 64.3  & 61.4   & 63.6  \\ \cline{2-9}
                                & \multirow{2}{*}{{R@10 $\uparrow$}}    & T2V & 63.4   & 57.6  & 63.3    & 73.0  & 61.8   & 71.4  \\
                                &                                       & V2T & 63.2   & 57.0  & 63.1    & 71.5  & 67.9   & 70.2  \\ \hline
\multirow{7}{*}{SSV2-label}     & GFLOPs $\downarrow$                   & (G) & 303.3  & 232.2 & 212.9   & 984.6 & 627.2  & 610.9 \\  \cline{2-9} 
                                & \multirow{2}{*}{{R@1 $\uparrow$}}     & T2V & 64.1   & 58.7  & 64.3    & 73.3  & 69.4   & 73.1  \\
                                &                                       & V2T & 63.8   & 61.7  & 63.3    & 71.5  & 70.3   & 71.1  \\ \cline{2-9}
                                & \multirow{2}{*}{{R@5 $\uparrow$ }}    & T2V & 88.2   & 84.9  & 88.0    & 93.7  & 92.1   & 93.3  \\
                                &                                       & V2T & 88.3   & 87.2  & 87.4    & 93.1  & 92.3   & 92.7  \\ \cline{2-9}
                                & \multirow{2}{*}{{R@10 $\uparrow$}}    & T2V & 92.7   & 90.8  & 92.6    & 96.6  & 95.8   & 96.6  \\
                                &                                       & V2T & 93.2   & 92.4  & 92.9    & 96.7  & 95.8   & 96.4  \\ \hline
\multirow{7}{*}{SSV2-Template}  & GFLOPs $\downarrow$                   & (G) & 303.3  & 241.4 & 203.7   & 984.6 & 627.2  & 572.9 \\  \cline{2-9} 
                                & \multirow{2}{*}{{R@1 $\uparrow$}}     & T2V & 87.9   & 85.1  & 87.4    & 90.8  & 88.5   & 90.2  \\
                                &                                       & V2T & 61.3   & 59.5  & 60.6    & 66.1  & 65.5   & 66.0  \\ \cline{2-9}
                                & \multirow{2}{*}{{R@5 $\uparrow$ }}    & T2V & 99.4   & 100.0 & 99.4    & 99.4  & 98.9   & 100.0 \\
                                &                                       & V2T & 88.5   & 87.9  & 87.4    & 92.3  & 91.4   & 91.7  \\ \cline{2-9}
                                & \multirow{2}{*}{{R@10 $\uparrow$}}    & T2V & 100.0  & 100.0 & 100.0   & 100.0 & 100.0  & 100.0 \\
                                &                                       & V2T & 93.7   & 93.1  & 92.6    & 95.6  & 95.5   & 95.8  \\ \hline
\thickhline
\end{tabular}
\vspace{-10pt}
\end{table*}

%% file: Supple/zero_shot_mean.tex
\begin{table*}[h]
\centering
\caption{
    \textbf{Comparison of zero-shot retrieval performance for four datasets}.
}
\label{sup:zero-shot}
\begin{tabular}{c|c|c|c|cc|c|cc}
\thickhline

\multirow{2}{*}{Dataset} & \multirow{2}{*}{\#Pairs} & \multirow{2}{*}{Metic} &  \multicolumn{3}{c|}{UMT-B} & \multicolumn{3}{c}{UMT-L} \\
 & & & \multicolumn{1}{c}{Base} & \multicolumn{1}{c}{ToMe} & \multicolumn{1}{c|}{Ours} & \multicolumn{1}{c}{Base} & \multicolumn{1}{c}{ToMe} & \multicolumn{1}{c}{Ours} \\ \hline

\multirow{12}{*}{MSVD} & \multirow{4}{*}{5M} & GFLOPs $\downarrow$   & 78.5 & 66.2 & \textbf{50.6} & 267.8 & 182.1 & \textbf{165.3} \\ \cline{3-9}
 &                      & R@1  $\uparrow$      & 47.4                 & 45.6 \scriptsize{(-1.8)} & \textbf{48.1} \scriptsize{(+0.7)} & 55.3 & 53.6 \scriptsize{(-1.7)} & \textbf{56.3} \scriptsize{(+1.0)} \\  
 &                      & R@5  $\uparrow$      & 72.2                 & 70.8 \scriptsize{(-1.4)} & \textbf{72.5} \scriptsize{(+0.3)} & 79.4 & 78.9 \scriptsize{(-0.5)} & \textbf{79.6} \scriptsize{(+0.2)} \\ 
 &                      & R@10 $\uparrow$      & 80.2                 & 79.8 \scriptsize{(-0.4)} & \textbf{80.4} \scriptsize{(+0.2)} & 85.9 & 85.1 \scriptsize{(-0.8)} & \textbf{86.1} \scriptsize{(+0.2)} \\ \cline{2-9}
 & \multirow{4}{*}{17M} & GFLOPs $\downarrow$  & 78.5 & 62.8 & \textbf{50.2} & 267.8 & 182.1 & \textbf{165.3} \\ \cline{3-9}
 &                      & R@1  $\uparrow$      & 52.0                 & 50.7 \scriptsize{(-1.3)}	& \textbf{53.3} \scriptsize{(+1.3)} & \textbf{62.7} & 61.0 \scriptsize{(-1.7)} & \textbf{62.7} \scriptsize{($\pm$0.0)} \\ 
 &                      & R@5  $\uparrow$      & 75.7                 & 74.8 \scriptsize{(-0.9)}	& \textbf{77.0} \scriptsize{(+1.3)} & 83.7 & 83.2 \scriptsize{(-0.5)} & \textbf{83.8} \scriptsize{(+0.1)} \\ 
 &                      & R@10 $\uparrow$      & 83.6                 & 83.0 \scriptsize{(-0.6)}	& \textbf{84.0} \scriptsize{(+0.4)} & 89.7 & 88.8 \scriptsize{(-0.9)} & \textbf{90.3} \scriptsize{(+0.6)} \\ \cline{2-9}
 & \multirow{4}{*}{25M} & GFLOPs $\downarrow$  & 78.5 & 66.2 & \textbf{50.6} & 267.8 & 182.1 & \textbf{153.6} \\ \cline{3-9}
 &                      & R@1  $\uparrow$      & 52.1                 & 50.1 \scriptsize{(-2.0)} & \textbf{53.6} \scriptsize{(+1.5)} & 61.8 & 60.3 \scriptsize{(-1.5)} & \textbf{62.5} \scriptsize{(+0.7)} \\ 
 &                      & R@5  $\uparrow$      & 77.1                 & 76.1 \scriptsize{(-1.0)} & \textbf{78.1} \scriptsize{(+1.0)} & 83.3 & 82.5 \scriptsize{(-0.8)} & \textbf{83.2} \scriptsize{(-0.1)} \\ 
 &                      & R@10 $\uparrow$      & 84.7                 & 84.1 \scriptsize{(-0.6)} & \textbf{85.4} \scriptsize{(+0.7)} & 88.8 & 88.3 \scriptsize{(-0.5)} & \textbf{88.8} \scriptsize{($\pm$0.0)} \\ \hline

 \multirow{12}{*}{MSRVTT}  & \multirow{4}{*}{5M} & GFLOPs $\downarrow$ & 78.5 & 66.2 & \textbf{50.2} & 267.8 & 182.1 & \textbf{164.8} \\ \cline{3-9}
 &                         & R@1  $\uparrow$     & 27.9  & 26.4 \scriptsize{(-1.5)} & \textbf{27.7} \scriptsize{(-0.2)}     & 31.8 & 30.4 \scriptsize{(-1.4)}	& \textbf{32.7} \scriptsize{(+0.9)} \\ 
 &                         & R@5  $\uparrow$     & 49.8  & 46.7 \scriptsize{(-3.1)} & \textbf{49.8} \scriptsize{($\pm$0.0)} & 54.7 & 52.4 \scriptsize{(-2.3)}	& \textbf{55.1} \scriptsize{(+0.4)} \\
 &                         & R@10 $\uparrow$     & 58.4  & 55.9 \scriptsize{(-2.5)} & \textbf{58.3} \scriptsize{(-0.1)}     & 64.2 & 62.9 \scriptsize{(-1.3)}	& \textbf{64.6} \scriptsize{(+0.4)} \\ \cline{2-9}
 & \multirow{4}{*}{17M}    & GFLOPs $\downarrow$ & 78.5  & 66.2                     & \textbf{50.6}                         & 267.8 & 182.1 & \textbf{163.0} \\ \cline{3-9}
 &                         & R@1  $\uparrow$     & 33.6  & 32.5 \scriptsize{(-1.1)} & \textbf{33.8} \scriptsize{(+0.2)}     & 40.6 & 37.6 \scriptsize{(-3.0)} & \textbf{39.9} \scriptsize{(-0.7)} \\
 &                         & R@5  $\uparrow$     & 56.4  & 54.1 \scriptsize{(-2.3)} & \textbf{57.0} \scriptsize{(+0.6)}     & 62.0 & 60.1 \scriptsize{(-1.9)} & \textbf{61.9} \scriptsize{(-0.1)} \\
 &                         & R@10 $\uparrow$     & 66.4  & 63.9 \scriptsize{(-2.5)} & \textbf{66.6} \scriptsize{(+0.2)}     & 71.4 & 68.9 \scriptsize{(-2.5)} & \textbf{70.9} \scriptsize{(-0.5)} \\ \cline{2-9}
 & \multirow{4}{*}{25M}    & GFLOPs $\downarrow$ & 78.5  & 66.2                     & \textbf{44.7}                         & 267.8 & 160.9 & \textbf{146.5} \\ \cline{3-9}
 &                         & R@1  $\uparrow$     & 32.8  & 31.1 \scriptsize{(-1.7)} & \textbf{33.3} \scriptsize{(+0.5)}     & 38.9 & 36.3 \scriptsize{(-2.6)} & \textbf{39.2} \scriptsize{(+0.3)} \\ 
 &                         & R@5  $\uparrow$     & 54.3  & 52.8 \scriptsize{(-1.5)} & \textbf{54.7} \scriptsize{(+0.4)}     & 61.1 & 59.0 \scriptsize{(-2.1)} & \textbf{61.2} \scriptsize{(+0.1)} \\ 
 &                         & R@10 $\uparrow$     & 32.8  & 31.1 \scriptsize{(-1.7)} & \textbf{33.3} \scriptsize{(+0.5)}     & 38.9 & 36.3 \scriptsize{(-2.6)} & \textbf{39.2} \scriptsize{(+0.3)} \\ \hline

 \multirow{12}{*}{DIDEMO} & \multirow{4}{*}{5M} & GFLOPs $\downarrow$ & 78.5 & 66.2 & \textbf{47.3} & 267.8 & 182.1 & \textbf{166.2} \\ \cline{3-9}
 &  & R@1  $\uparrow$ & 35.8 & 29.7 \scriptsize{(-6.1)} & \textbf{35.9} \scriptsize{(+0.1)} & 35.1 & 33.7 \scriptsize{(-1.4)}  & \textbf{36.6} \scriptsize{(+1.5)} \\ 
 &  & R@5  $\uparrow$ & 61.9 & 55.5 \scriptsize{(-6.4)} & \textbf{60.6} \scriptsize{(-1.3)} & 60.2 & 60.0 \scriptsize{(-0.2)} & \textbf{62.1} \scriptsize{(+1.9)} \\ 
 &  & R@10 $\uparrow$ & 70.1 & 65.1 \scriptsize{(-5.0)} & \textbf{69.0} \scriptsize{(-1.1)} & 68.7 & 69.5 \scriptsize{(+0.8)}  & \textbf{71.0} \scriptsize{(+2.3)} \\ \cline{2-9}
 & \multirow{4}{*}{17M}    & GFLOPs $\downarrow$ & 78.5 & 62.8 & \textbf{44.5} & 267.8 & 182.1 & \textbf{166.7} \\ \cline{3-9}
 &  & R@1  $\uparrow$ & 41.1 & 38.6 \scriptsize{(-2.5)} & \textbf{42.3} \scriptsize{(+1.2)} & 46.5 & 45.4 \scriptsize{(-1.0)} & \textbf{48.7} \scriptsize{(+2.2)} \\ 
 &  & R@5  $\uparrow$ & 66.7 & 64.7 \scriptsize{(-2.0)} & \textbf{68.4} \scriptsize{(+1.7)} & 71.1 & 71.8 \scriptsize{(+0.7)}  & \textbf{74.3} \scriptsize{(+3.2)} \\ 
 &  & R@10 $\uparrow$ & 75.4 & 73.8 \scriptsize{(-1.6)} & \textbf{77.0} \scriptsize{(+1.6)} & 79.2 & 80.1 \scriptsize{(+0.9)}  & \textbf{81.3} \scriptsize{(+2.1)} \\ \cline{2-9}
 & \multirow{4}{*}{25M}    & GFLOPs $\downarrow$ & 78.5 & 62.8 & \textbf{48.2} & 267.8 & 182.1 & \textbf{162.5} \\ \cline{3-9}
 &  & R@1  $\uparrow$ & 41.0 & 38.6 \scriptsize{(-2.4)} & \textbf{44.0} \scriptsize{(+3.0)} & 49.3 & 47.9 \scriptsize{(-1.3)} & \textbf{52.0} \scriptsize{(+2.8)} \\ 
 &  & R@5  $\uparrow$ & 66.6 & 64.5 \scriptsize{(-2.1)} & \textbf{68.9} \scriptsize{(+2.3)} & 73.9 & 72.6 \scriptsize{(-1.3)} & \textbf{74.9} \scriptsize{(+1.1)} \\ 
 &  & R@10 $\uparrow$ & 75.8 & 73.0 \scriptsize{(-2.8)} & \textbf{77.7} \scriptsize{(+1.9)} & 80.2 & 81.1 \scriptsize{(+0.9)}  & \textbf{82.4} \scriptsize{(+2.2)} \\ \hline

 \multirow{12}{*}{ANET} & \multirow{4}{*}{5M} & GFLOPs $\downarrow$ & 78.5 & 66.2 & \textbf{49.0} & 267.8 & 182.1 & \textbf{162.0} \\ \cline{3-9}
 &  & R@1  $\uparrow$ & 27.1 & 24.9 \scriptsize{(-2.2)} & \textbf{28.1} \scriptsize{(+1.0)} & 31.0 & 29.7 \scriptsize{(-1.3)} & \textbf{31.3} \scriptsize{(+0.3)} \\ 
 &  & R@5  $\uparrow$ & 51.6 & 49.1 \scriptsize{(-2.5)} & \textbf{52.4} \scriptsize{(+0.8)} & 59.7 & 57.9 \scriptsize{(-1.8)} & \textbf{59.6} \scriptsize{(-0.1)} \\ 
 &  & R@10 $\uparrow$ & 63.0 & 61.0 \scriptsize{(-2.0)} & \textbf{64.1} \scriptsize{(+1.1)} & 71.7 & 70.2 \scriptsize{(-1.5)} & \textbf{71.6} \scriptsize{(-0.1)} \\ \cline{2-9}
 & \multirow{4}{*}{17M}     & GFLOPs $\downarrow$ & 78.5 & 62.8 & \textbf{45.0} & 267.8 & 182.1 & \textbf{151.7} \\ \cline{3-9}
 &  & R@1  $\uparrow$ & 32.7 & 31.7 \scriptsize{(-1.0)} & \textbf{33.9} \scriptsize{(+1.2)} & 41.8 & 40.3 \scriptsize{(-1.5)} & \textbf{41.7} \scriptsize{(-0.1)} \\ 
 &  & R@5  $\uparrow$ & 57.7 & 57.4 \scriptsize{(-0.3)} & \textbf{59.6} \scriptsize{(+1.9)} & 68.6 & 67.1 \scriptsize{(-1.5)} & \textbf{68.8} \scriptsize{(+0.2)} \\ 
 &  & R@10 $\uparrow$ & 69.2 & 68.9 \scriptsize{(-0.3)} & \textbf{70.7} \scriptsize{(+1.5)} & 79.2 & 78.2 \scriptsize{(-1.0)} & \textbf{79.1} \scriptsize{(-0.1)} \\ \cline{2-9}
 & \multirow{4}{*}{25M}     & GFLOPs $\downarrow$ & 78.5 & 62.8 & \textbf{51.0} & 267.8 & 182.1 & \textbf{152.7} \\ \cline{3-9}
 &  & R@1  $\uparrow$ & 34.2 & 32.6 \scriptsize{(-1.6)} & \textbf{35.7} \scriptsize{(+1.5)} & 40.7 & 41.0  \scriptsize{(+0.3)} & \textbf{41.8} \scriptsize{(+1.1)}\\ 
 &  & R@5  $\uparrow$ & 59.1 & 58.1 \scriptsize{(-1.0)} & \textbf{60.9} \scriptsize{(+1.8)} & 67.9 & 67.8 \scriptsize{(-0.1)} & \textbf{68.4} \scriptsize{(+0.5)}\\ 
 &  & R@10 $\uparrow$ & 70.5 & 69.5 \scriptsize{(-1.0)} & \textbf{72.0} \scriptsize{(+1.5)} & 79.3 & 78.4 \scriptsize{(-0.9)} & \textbf{79.0} \scriptsize{(-0.3)}\\

\thickhline
\end{tabular}
\end{table*}

%% file: Supple/zero_shot_1.tex
\renewcommand{\arraystretch}{1.0}
\begin{table*}[t]
\centering
\caption{
    \textbf{Detailed zero-shot retrieval performance on MSVD.}
}

\label{sup:zs_msvd}
\begin{tabular}{c|c|c|c|c|cc|c|cc}
\thickhline
\multirow{2}{*}{Dataset} & \multirow{2}{*}{Pairs} & \multirow{2}{*}{Metic} & \multirow{2}{*}{Type} & \multicolumn{3}{c}{UMT-B} & \multicolumn{3}{c}{UMT-L} \\
 & & & & \multicolumn{1}{c}{Base} & \multicolumn{1}{c}{ToMe} & \multicolumn{1}{c}{Ours} & \multicolumn{1}{c}{Base} & \multicolumn{1}{c}{ToMe} & \multicolumn{1}{c}{Ours} \\ \hline

\multirow{21}{*}{MSVD} & \multirow{7}{*}{5M} & FLOPS $\downarrow$ & (G)& 78.5 & 66.2 & 50.6 & 267.8 & 182.1 & 165.3 \\ \cline{3-10}
 & & \multirow{2}{*}{R@1 $\uparrow$} & T2V & 36.2 & 33.6 & 37.0 & 44.4 & 42.0 & 44.3 \\
 & & & V2T & 58.5 & 57.5 & 59.1 & 66.1 & 65.2 & 68.2 \\ \cline{4-10}
 \cline{3-10}
 & & \multirow{2}{*}{R@5 $\uparrow$} & T2V & 65.7 & 64.3 & 66.1 & 73.3 & 72.0 & 73.2 \\
 & & & V2T & 78.7 & 77.3 & 78.8 & 85.5 & 85.8 & 86.0 \\
 \cline{3-10}
 & & \multirow{2}{*}{R@10 $\uparrow$} & T2V & 76.1 & 74.8 & 76.4 & 82.4 & 80.5 & 82.0 \\
 & & & V2T & 84.3 & 84.8 & 84.3 & 89.4 & 89.6 & 90.2 \\
 \cline{2-10}
 & \multirow{7}{*}{17M} & FLOPS $\downarrow$ & (G)& 78.5 & 62.8 & 50.2 & 267.8 & 182.1 & 165.3\\ \cline{3-10}
 & & \multirow{2}{*}{R@1 $\uparrow$} & T2V & 41.4 & 38.6 & 42.2 & 49.9 & 48.1 & 50.0 \\
 & & & V2T & 62.5 & 62.8 & 64.3 & 75.4 & 73.9 & 75.4 \\
 \cline{3-10}
 & & \multirow{2}{*}{R@5 $\uparrow$} & T2V & 70.6 & 69.0 & 71.3 & 77.7 & 76.6 & 77.6 \\
 & & & V2T & 80.8 & 80.5 & 82.7 & 89.6 & 89.7 & 90.0 \\
 \cline{3-10}
 & & \multirow{2}{*}{R@10 $\uparrow$} & T2V & 80.1 & 78.9 & 80.5 & 85.3 & 84.3 & 85.5 \\
 & & & V2T & 87.0 & 87.0 & 87.5 & 94.0 & 93.3 & 95.1 \\
 \cline{2-10}
 & \multirow{7}{*}{25M} & FLOPS $\downarrow$ & (G)& 78.5 & 66.2 & 50.6 & 267.8 & 182.1 & 153.6 \\ \cline{3-10}
 & & \multirow{2}{*}{R@1 $\uparrow$} & T2V & 42.3 & 39.0 & 42.7 & 49.0 & 48.1 & 49.3 \\
 & & & V2T & 61.9 & 61.2 & 64.5 & 74.5 & 72.4 & 75.7 \\
 \cline{3-10}
 & & \multirow{2}{*}{R@5 $\uparrow$} & T2V & 71.7 & 69.6 & 72.1 & 76.9 & 75.4 & 77.0 \\
 & & & V2T & 82.5 & 82.5 & 84.0 & 89.7 & 89.6 & 89.4 \\
 \cline{3-10}
 & & \multirow{2}{*}{R@10 $\uparrow$} & T2V & 80.8 & 79.4 & 81.3 & 84.7 & 83.2 & 84.2 \\
 & & & V2T & 88.5 & 88.8 & 89.4 & 92.8 & 93.3 & 93.3 \\

\thickhline
\end{tabular}
\end{table*}

\begin{table*}[t]
\centering
\caption{
    \textbf{Detailed zero-shot retrieval performance on MSRVTT.}
}

\label{sup:zs_msrvtt}
\begin{tabular}{c|c|c|c|c|cc|c|cc}
\thickhline
\multirow{2}{*}{Dataset} & \multirow{2}{*}{Pairs} & \multirow{2}{*}{Metic} & \multirow{2}{*}{Type} & \multicolumn{3}{c}{UMT-B} & \multicolumn{3}{c}{UMT-L} \\
 & & & & \multicolumn{1}{c}{Base} & \multicolumn{1}{c}{ToMe} & \multicolumn{1}{c}{Ours} & \multicolumn{1}{c}{Base} & \multicolumn{1}{c}{ToMe} & \multicolumn{1}{c}{Ours} \\ \hline
\multirow{21}{*}{MSRVTT} & \multirow{7}{*}{5M} & FLOPS $\downarrow$ & (G)& 78.5 & 66.2 & 50.2 & 267.8 & 182.1 & 164.8 \\ \cline{3-10}
 & & \multirow{2}{*}{R@1 $\uparrow$} & T2V & 29.6 & 27.8 & 29.9 & 33.3 & 31.4 & 34.0 \\
 & & & V2T & 26.2 & 25.0 & 25.4 & 30.2 & 29.4 & 31.3 \\ 
 \cline{3-10}
 & & \multirow{2}{*}{R@5 $\uparrow$} & T2V & 52.8 & 48.4 & 52.3 & 58.1 & 53.3 & 57.1 \\
 & & & V2T & 46.7 & 44.9 & 47.3 & 51.3 & 51.4 & 53.0 \\
 \cline{3-10}
 & & \multirow{2}{*}{R@10 $\uparrow$} & T2V & 61.9 & 57.4 & 61.6 & 66.7 & 64.4 & 66.1 \\
 & & & V2T & 54.9 & 54.3 & 55.0 & 61.6 & 61.4 & 63.1 \\
 \cline{2-10}
 & \multirow{7}{*}{17M} & FLOPS $\downarrow$ & (G)& 78.5 & 66.2 & 50.6 & 267.8 & 182.1 & 163.0 \\ \cline{3-10}
 & & \multirow{2}{*}{R@1 $\uparrow$} & T2V & 35.5 & 32.5 & 36.5 & 42.6 & 39.7 & 42.1 \\
 & & & V2T & 31.6 & 32.4 & 31.0 & 38.6 & 35.5 & 37.7 \\
 \cline{3-10}
 & & \multirow{2}{*}{R@5 $\uparrow$} & T2V & 59.3 & 55.8 & 59.6 & 64.4 & 62.0 & 63.9 \\
 & & & V2T & 53.5 & 52.3 & 54.4 & 59.6 & 58.1 & 59.8 \\
 \cline{3-10}
 & & \multirow{2}{*}{R@10 $\uparrow$} & T2V & 68.6 & 64.1 & 68.9 & 73.1 & 70.5 & 72.4 \\
 & & & V2T & 64.1 & 63.7 & 64.2 & 69.6 & 67.3 & 69.4 \\
 \cline{2-10}
 & \multirow{7}{*}{25M} & FLOPS $\downarrow$ & (G)& 78.5 & 66.2 & 44.7 & 267.8 & 160.9 & 146.5 \\ \cline{3-10}
 & & \multirow{2}{*}{R@1 $\uparrow$} & T2V & 35.2 & 31.7 & 35.3 & 40.7 & 37.9 & 41.0 \\
 & & & V2T & 30.3 & 30.4 & 31.3 & 37.1 & 34.6 & 37.4 \\
 \cline{3-10}
 & & \multirow{2}{*}{R@5 $\uparrow$} & T2V & 57.8 & 53.9 & 57.4 & 63.4 & 61.4 & 63.6 \\
 & & & V2T & 50.7 & 51.7 & 52.0 & 58.7 & 56.6 & 58.8 \\
 \cline{3-10}
 & & \multirow{2}{*}{R@10 $\uparrow$} & T2V & 66.0 & 61.8 & 65.6 & 71.8 & 69.0 & 71.5 \\
 & & & V2T & 61.4 & 62.2 & 62.8 & 68.9 & 66.2 & 68.6 \\
\thickhline
\end{tabular}
\end{table*}

%% file: Supple/zero_shot_2.tex
\begin{table*}[t]
\centering
\caption{
    \textbf{Detailed zero-shot retrieval performance on DiDeMo}
}
\label{sup:zs_didemo}
\begin{tabular}{c|c|c|c|c|cc|c|cc}
\thickhline
\multirow{2}{*}{Dataset} & \multirow{2}{*}{\#Pairs} & \multirow{2}{*}{Metic} & \multirow{2}{*}{Type} & \multicolumn{3}{c|}{UMT-B} & \multicolumn{3}{c}{UMT-L} \\
                         &                      &                                  &     & Base & ToMe & Ours & Base  & ToMe  & Ours           \\ \hline
\multirow{21}{*}{DiDeMo} & \multirow{7}{*}{5M}  & FLOPs $\downarrow$               & (G) & 78.5 & 66.2 & 47.3 & 267.8 & 182.1 & 166.2 \\ \cline{3-10}
                         &                      & \multirow{2}{*}{R@1 $\uparrow$}  & T2V & 35.7 & 28.0 & 35.6 & 34.0  & 32.1  & 35.9  \\
                         &                      &                                  & V2T & 35.8 & 31.4 & 36.1 & 36.2  & 35.3  & 37.2  \\ \cline{3-10}
                         &                      & \multirow{2}{*}{R@5 $\uparrow$}  & T2V & 62.2 & 52.9 & 61.2 & 60.4  & 57.7  & 61.0  \\
                         &                      &                                  & V2T & 61.6 & 58.1 & 60.0 & 60.0  & 62.3  & 63.2  \\ \cline{3-10}
                         &                      & \multirow{2}{*}{R@10 $\uparrow$} & T2V & 70.6 & 62.3 & 69.3 & 68.7  & 67.9  & 70.2  \\
                         &                      &                                  & V2T & 69.6 & 67.9 & 68.7 & 68.6  & 71.2  & 71.9  \\ \cline{2-10}
                         & \multirow{7}{*}{17M} & FLOPs $\downarrow$               & (G) & 78.5 & 62.8 & 44.5 & 267.8 & 182.1 & 166.7 \\ \cline{3-10}
                         &                      & \multirow{2}{*}{R@1 $\uparrow$}  & T2V & 41.9 & 36.9 & 42.6 & 46.4  & 44.8  & 49.7  \\
                         &                      &                                  & V2T & 40.3 & 40.3 & 42.0 & 46.5  & 46.0  & 47.6  \\ \cline{3-10}
                         &                      & \multirow{2}{*}{R@5 $\uparrow$}  & T2V & 66.7 & 63.0 & 68.6 & 70.0  & 72.3  & 73.9  \\
                         &                      &                                  & V2T & 66.6 & 66.4 & 68.3 & 72.2  & 71.3  & 74.8  \\ \cline{3-10}
                         &                      & \multirow{2}{*}{R@10 $\uparrow$} & T2V & 75.0 & 71.9 & 77.2 & 78.8  & 80.2  & 80.9  \\
                         &                      &                                  & V2T & 75.8 & 75.8 & 76.8 & 79.5  & 79.9  & 81.6  \\ \cline{2-10}
                         & \multirow{7}{*}{25M} & FLOPs $\downarrow$               & (G) & 78.5 & 62.8 & 48.2 & 267.8 & 160.9 & 162.5 \\ \cline{3-10}
                         &                      & \multirow{2}{*}{R@1 $\uparrow$}  & T2V & 41.2 & 35.6 & 44.6 & 48.6  & 47.9  & 52.0  \\
                         &                      &                                  & V2T & 40.8 & 41.5 & 43.4 & 49.9  & 48.4  & 52.0  \\ \cline{3-10}
                         &                      & \multirow{2}{*}{R@5 $\uparrow$}  & T2V & 65.4 & 62.0 & 68.9 & 72.9  & 72.9  & 74.0  \\
                         &                      &                                  & V2T & 67.7 & 67.0 & 68.9 & 74.8  & 74.1  & 75.9  \\ \cline{3-10}
                         &                      & \multirow{2}{*}{R@10 $\uparrow$} & T2V & 74.9 & 70.2 & 76.8 & 79.0  & 80.1  & 81.0  \\
                         &                      &                                  & V2T & 76.7 & 75.9 & 78.6 & 81.4  & 81.6  & 83.8  \\
\thickhline
\end{tabular}
\end{table*}

\begin{table*}[t]
\centering
\caption{
    \textbf{Detailed zero-shot retrieval performance on Activitynet.}
}
\label{sup:zs_anet}
\begin{tabular}{c|c|c|c|c|cc|c|cc}
\thickhline
\multirow{2}{*}{Dataset} & \multirow{2}{*}{\#Pairs} & \multirow{2}{*}{Metic} & \multirow{2}{*}{Type} & \multicolumn{3}{c|}{UMT-B} & \multicolumn{3}{c}{UMT-L} \\
 &  &  &  & Base & ToMe & Ours & Base & ToMe & Ours \\ \hline
\multirow{21}{*}{ANET} & \multirow{7}{*}{5M} & FLOPs $\downarrow$ & (G) & 78.5 & 66.2 & 49.0 & 267.8 & 182.1 & 162.0 \\ \cline{3-10}
 &  & \multirow{2}{*}{R@1 $\uparrow$} & T2V & 28.3 & 24.7 & 29.3 & 31.9 & 29.4 & 31.4 \\
 &  &  & V2T & 25.9 & 25.1 & 26.8 & 30.0 & 30.1 & 31.1 \\
 \cline{3-10}
 &  & \multirow{2}{*}{R@5 $\uparrow$} & T2V & 53.0 & 48.9 & 53.2 & 60.2 & 57.4 & 59.6 \\
 &  &  & V2T & 50.2 & 49.4 & 51.6 & 59.1 & 58.5 & 59.6 \\
 \cline{3-10}
 &  & \multirow{2}{*}{R@10 $\uparrow$} & T2V & 64.2 & 60.5 & 64.3 & 72.0 & 69.6 & 71.5 \\
 &  &  & V2T & 61.7 & 61.4 & 64.0 & 71.3 & 70.9 & 71.7 \\
 \cline{2-10}
 & \multirow{7}{*}{17M} & FLOPs $\downarrow$ & (G) & 78.5 & 62.8 & 45.0 & 267.8 & 182.1 & 151.7 \\ \cline{3-10}
 &  & \multirow{2}{*}{R@1 $\uparrow$} & T2V & 33.8 & 31.5 & 34.6 & 42.8 & 40.1 & 42.1 \\
 &  &  & V2T & 31.6 & 32.0 & 33.2 & 40.7 & 40.4 & 41.2 \\
 \cline{3-10}
 &  & \multirow{2}{*}{R@5 $\uparrow$} & T2V & 59.1 & 57.4 & 60.7 & 69.6 & 67.0 & 69.4 \\
 &  &  & V2T & 56.2 & 57.5 & 58.5 & 67.6 & 67.2 & 68.2 \\
 \cline{3-10}
 &  & \multirow{2}{*}{R@10 $\uparrow$} & T2V & 70.4 & 68.5 & 71.1 & 79.8 & 77.9 & 79.2 \\
 &  &  & V2T & 67.9 & 69.2 & 70.3 & 78.6 & 78.5 & 79.1 \\
 \cline{2-10}
 & \multirow{7}{*}{25M} & FLOPs $\downarrow$ & (G) & 78.5 & 62.8 & 51.0 & 267.8 & 182.1 & 152.7 \\ \cline{3-10}
 &  & \multirow{2}{*}{R@1 $\uparrow$} & T2V & 35.5 & 32.6 & 36.4 & 41.9 & 41.6 & 42.8 \\
 &  &  & V2T & 32.8 & 32.6 & 34.9 & 39.5 & 40.4 & 40.8 \\
 \cline{3-10}
 &  & \multirow{2}{*}{R@5 $\uparrow$} & T2V & 60.6 & 57.8 & 61.3 & 68.6 & 68.1 & 68.8 \\
 &  &  & V2T & 57.6 & 58.5 & 60.5 & 67.6 & 67.5 & 67.9 \\
 \cline{3-10}
 &  & \multirow{2}{*}{R@10 $\uparrow$} & T2V & 71.8 & 69.0 & 72.4 & 79.6 & 78.7 & 79.6 \\
 &  &  & V2T & 69.2 & 70.1 & 71.6 & 78.4 & 78.2 & 78.5 \\

 \thickhline
\end{tabular}
\end{table*}